\documentclass[sigconf,nonacm]{acmart}
\AtBeginDocument{%
  \providecommand\BibTeX{{%
    \normalfont B\kern-0.5em{\scshape i\kern-0.25em b}\kern-0.8em\TeX}}}

\setcopyright{acmcopyright}
\copyrightyear{2022}
\acmYear{2022}
\acmDOI{XX.XXXX/XXXXXXX.XXXXXXX}

\acmConference[PACT '22]{PACT '22:  International Conference on
Parallel Architectures and Compilation Techniques (PACT)}{October 10--12, 2022}{Chicago,IL}
\acmBooktitle{PACT '22: International Conference on
Parallel Architectures and Compilation Techniques (PACT), October 10--12, 2022, Chicago, IL}
\acmPrice{15.00}
\acmISBN{XXX-X-XXXX-XXXX-X/XX/XX}

\usepackage{tikz}

\usepackage{subcaption} % for subfigures

\usepackage{algorithm,refcount} % for algorithms
\usepackage[noend]{algpseudocode} % for algorithms

\usepackage{bm} % bold math

\usepackage[inline]{enumitem} % inline enumeration

\usepackage{multirow} % multirow table cols

\usepackage{pifont} % xmark and cmark

\usepackage{makecell} % linebreaks in tables
\usepackage{lipsum}  

\usepackage{tabularx}
\newcolumntype{C}{>{\centering\arraybackslash}X}

\newcolumntype{Q}[1]{>{\centering\arraybackslash}p{#1}}
\newcolumntype{M}[1]{>{\centering\arraybackslash}m{#1}}
\usepackage{threeparttable}
\begin{document}

%**********************************************************************************************

%%
%% The "title" command has an optional parameter,
%% allowing the author to define a "short title" to be used in page headers.

%\title{Transfer-Tuning: Exploiting Workload Similarity for Efficient Tensor Program Code Generation}
\title{Transfer-Tuning: Reusing Auto-Schedules for Efficient Tensor Program Code Generation}

%%
%% The "author" command and its associated commands are used to define
%% the authors and their affiliations.
%% Of note is the shared affiliation of the first two authors, and the
%% "authornote" and "authornotemark" commands
%% used to denote shared contribution to the research.
% \author{Perry Gibson}
% %\authornote{Both authors contributed equally to this research.}
% \email{p.gibson.2@research.gla.ac.uk}
% \orcid{0000-0003-3370-0698}
% \author{Jose Cano}
% \email{Jose.CanoReyes@glasgow.ac.uk}
% \orcid{0000-0002-2243-389X}
% \affiliation{%
%   \institution{University of Glasgow}
%   \streetaddress{P.O. Box 1212}
%   \city{Dublin}
%   \state{Ohio}
%   \country{USA}
%   \postcode{43017-6221}
% }
\author{Perry Gibson, Jos\'e Cano}
\orcid{0000-0003-3370-0698}
\orcid{0000-0002-2243-389X}
\affiliation{%
  \institution{School of Computing Science, University of Glasgow, Scotland \country{UK}}
  }
%\email{p.gibson.2@research.gla.ac.uk, Jose.CanoReyes@glasgow.ac.uk}

% \affiliation{%
%   \institution{University of Glasgow}}

%%
%% By default, the full list of authors will be used in the page
%% headers. Often, this list is too long, and will overlap
%% other information printed in the page headers. This command allows
%% the author to define a more concise list
%% of authors' names for this purpose.
\renewcommand{\shortauthors}{Gibson and Cano}

%**********************************************************************************************

%%
%% The abstract is a short summary of the work to be presented in the
%% article.
\begin{abstract}
%  under 300 words.

Auto-scheduling for tensor programs is a process where a search algorithm automatically explores candidate schedules (program transformations) for a given program on a target hardware platform to improve its performance.
However this can be a very time consuming process depending on the complexity of the tensor program and the capacity of the target device, with often many thousands of program variants being explored.
To address this, in this paper we introduce the idea of \emph{transfer-tuning}, a novel approach to identify and reuse auto-schedules between tensor programs.
We demonstrate this concept using Deep Neural Networks (DNNs), taking sets of auto-schedules from pre-tuned DNNs and using them to reduce the inference time of a new DNN.
We compare transfer-tuning against the state-of-the-art Ansor auto-scheduler, defining the maximum possible speedup for a given DNN model as what Ansor achieves using its recommended full tuning time. On a server-class CPU and across 11 widely used DNN models, we observe that transfer-tuning achieves up to $88.41\%$ ($49.13\%$ on average) of this maximum speedup, while Ansor requires $6.5\times$ more search time on average to match it.
We also evaluate transfer-tuning on a constrained edge CPU and observe that the differences in search time are exacerbated, with Ansor requiring $10.8\times$ more time on average to match transfer-tuning's speedup, which further demonstrates its value.
Our code is available at \url{https://github.com/gicLAB/transfer-tuning}.

\end{abstract}


%%
%% Keywords. The author(s) should pick words that accurately describe
%% the work being presented. Separate the keywords with commas.

%%
%% This command processes the author and affiliation and title
%% information and builds the first part of the formatted document.
\maketitle

\textbf{ACM Reference Format:}\\
Perry Gibson, José Cano. 2022. Transfer-Tuning: Reusing Auto-Schedules
for Efficient Tensor Program Code Generation. In PACT ’22: International
Conference on Parallel Architectures and Compilation Techniques (PACT),
October 10–12, 2022, Chicago, IL. ACM, New York, NY, USA, 12 pages. \url{https:
//doi.org/XX.XXXX/XXXXXXX.XXXXXXX}

\newpage
%*********************************************************************************************

\section{Introduction}
\label{sec:intro}

Computationally expensive tensor programs such as Deep Neural Networks (DNNs) have broad applications across fields such as computer vision~\cite{he2016deep,denseNet2017,7298965,yolo2016}, natural language processing~\cite{cnnsentence2014,Kalchbrenner14aconvolutional,devlin2019a}, scientific computing~\cite{zhang2019a,kutz2017}, and many more.
To achieve high performance on these DNN models, a range of efficient hand-tuned kernel libraries such as OpenBLAS~\cite{openblas} and oneDNN~\cite{onednn} for CPUs and cuDNN~\cite{chetlur2014} for NVIDIA GPUs have been developed to accelerate the performance of common operations.
However, these libraries require a great engineering effort to be optimized, do not necessarily see performance portability to new hardware architectures, and optimize for common cases often leaving novel operations with poor performance, such as capsule networks~\cite{sabour2017a}.
Reliance on hand-tuned kernel libraries can reduce the speed of adoption of new solutions~\cite{barham2019}, since communities in both academia and industry must wait for optimized operations to be developed by kernel library contributors.
Schedule based approaches such as Apache TVM~\cite{chen2018d} and Halide~\cite{ragan-kelley2017} can reduce some of these barriers by decoupling the high level description of the target algorithm from its platform specific optimizations.
This can make it easier to port algorithms to new systems, although still requires domain expertise to write optimized schedules.

%Ansor~\cite{zheng2020} is an auto-scheduling system which extends TVM and approaches the problem by iteratively exploring schedules to produce more efficient code for a given program and target hardware device. This process automatically generates schedules, compared to the hand-tuned schedules and kernels used by TVM, cuDNN, etc.
Ansor~\cite{zheng2020} is an \emph{auto-scheduling} system which extends TVM by automatically generating schedules for a given tensor program and target hardware device, compared to the hand-tuned schedules and kernels used by TVM, cuDNN, etc.
The approach can produce state-of-the-art inference time performance on a range of platforms and programs, and in particular can show improvement compared to existing approaches on novel operations such as capsule 2D convolution~\cite{sabour2017a}.
Ansor splits the computation graph into a set of kernels, where kernels are loop nests of easily fused operations such as convolutional layers and their complementary activation function, which are tuned individually for a given hardware device to produce their corresponding auto-schedules.

\begin{figure}[!t]
 \centering
 \includegraphics[width=0.99\columnwidth]{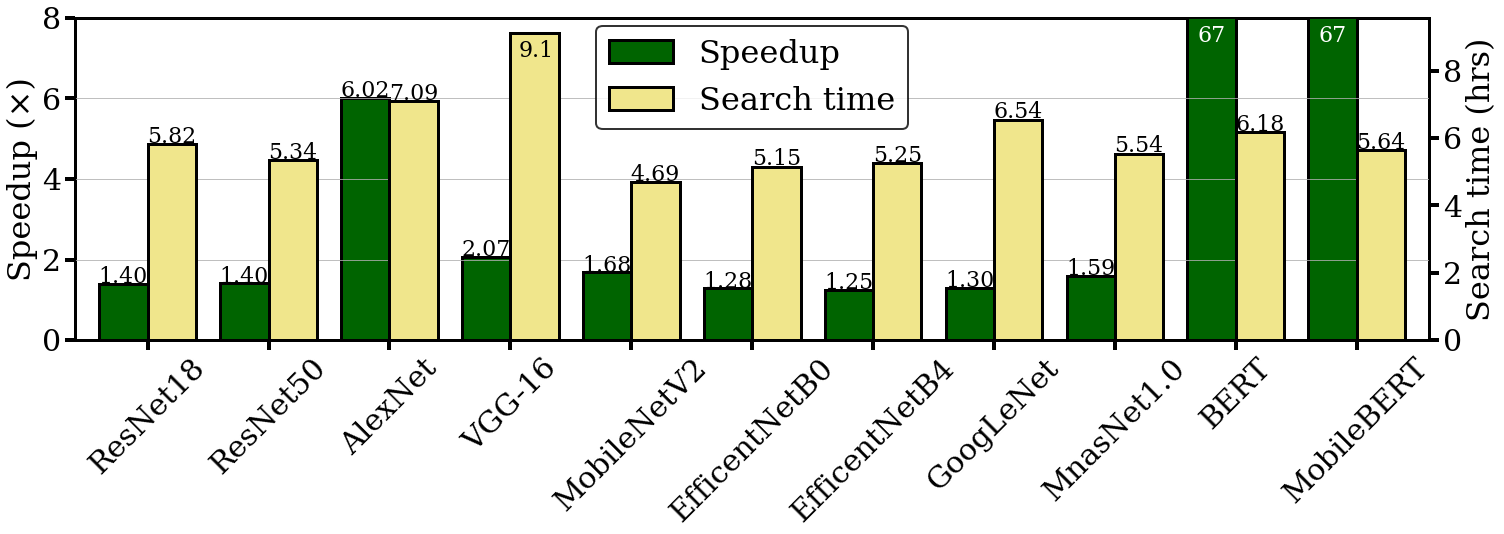}
 %\caption{\label{fig:motivation_time} Auto-scheduling searchhave added some time and inference time speedup using Ansor for an Intel Xeon E5-2620 with up to 20000 program variants.}
 \caption{\label{fig:motivation_time} Inference time speedup and auto-scheduling search time when running Ansor on an Intel Xeon E5-2620.}
\end{figure}

However, this tuning process can be very time consuming and specific to a given kernel of a given size. 
For example, if we tune a convolutional layer of some size and we are presented with a new convolutional layer of a different size, we must tune the new layer from scratch.
Figure~\ref{fig:motivation_time} shows the tuning time of the Ansor auto-scheduler for a number of widely used DNN models on a common server-class CPU, an Intel Xeon E5-2620.
We observe that the maximum speedup varies between models and the search time can take several hours, which is a large upfront cost.
%Note that 20,000 schedule variants is an arbitrary value, but is recommended by the TVM documentation as being generally sufficient to find the best performance.
If a range of applications are to be deployed on a given platform, this upfront cost may be further exacerbated and make the potential performance improvements of auto-scheduling impractical to achieve.
% \footnote{Note that Ansor auto-scheduling can be performed for both GPU and CPU hardware targets, however it currently requires a CPU to generate the schedule variants.}

To reduce tuning time we could sacrifice potential performance improvements by stopping early or only tuning a subset of the kernels.
However Ansor gets its improvements by evaluating a vast array of possible schedules, and both early stopping and tuning a subset of kernels may miss performant schedules.
Our main observation is that many applications, such as DNNs, feature high similarity between kernels in terms of the types of operations they compute and the sizes of their tensors.
For example, most DNNs contain a limited set of operations such as convolutional, dense, and pooling layers.
The DNN models in Figure~\ref{fig:motivation_time} contain $22$ unique kernel types (which we call \emph{kernel classes}), with every model having at least $1$ kernel class in common with every other model, and often many more.
Thus, if we have already found performant schedules for some tensor program, perhaps we could reuse this information on other tensor programs which contain similar kernels.
% on the target hardware platform 

In this paper we introduce \emph{transfer-tuning}, a novel approach which can improve execution performance for a given tensor program with reduced tuning time.
%when tuning time is limited.
Transfer-tuning exploits the similarity between kernels containing the same operations with varying data sizes, such that we can reuse schedules from other tensor programs.
Therefore, we can achieve performance improvements while reducing the costs associated with auto-scheduling.
Transfer-tuning's main value comes in use-cases where tensor program deployment requires performance efficiency but has reduced resources to perform costly auto-scheduling.

The contributions of this paper include the following:

\begin{itemize}[topsep=2.5pt]
\setlength\itemsep{2.5pt}
    \item We introduce \emph{transfer-tuning}, a new approach to reduce the costs of auto-scheduling based on reusing auto-schedules between tensor programs.

    \item We discuss in detail the key components of transfer-tuning such as kernel classes and a model selection heuristic, and how it is enabled by the features of the compute schedule programming paradigm. We also demonstrate the principles with an example using auto-schedules of two GEMM kernels of different sizes with each other, and tune the ResNet18 model using schedules from ResNet50.

    \item We evaluate transfer-tuning for $11$ representative DNN models obtaining a maximum speedup over untuned models of between $1.13\times$ and $59.4\times$, and compare it against Ansor which requires over $6.5\times$ as much search time to match our performance on average.
    We also evaluate transfer-tuning on a Raspberry Pi 4, a common constrained edge device, and observe that the differences in search time are exacerbated, with Ansor requiring over $10.8\times$ as much time to match our speedup on average.

    \item We present an alternative view on transfer-tuning, where the DNN model is the same but the input data size changes.
    
    % \color{red}
    % \item We explore an alternative formulation of transfer-tuning mixing schedules from several models, further demonstrating the flexibility of the approach, while exposing a direction of future work around across-kernel interactions.
    % \color{black}
\end{itemize}

% The rest of the paper is organized as follows.
% We provide the relevant background about the compute schedules paradigm, auto-tuning, and auto-scheduling in Section~\ref{sec:background}.
% In Section~\ref{sec:contr} we motivate, discuss the details, and provide an illustrative example of transfer-tuning.
% Section~\ref{sec:evaluation} demonstrates transfer-tuning evaluating 9 common DNN models, focusing on across-model transfer-tuning.
% A discussion of how transfer-tuning can be expanded is discussed in Section~\ref{sec:discusion}.
% In Section~\ref{sec:related_work} we discuss related work, and finally we conclude the paper in Section~\ref{sec:conclusion} briefly discussing potential areas for future work.

%*********************************************************************************************

\section{Background}%Schedules and auto-scheduling}
\label{sec:background}

Compute schedules is a programming paradigm which decouples the high-level description of an algorithm from the description of how it should be optimized for a given hardware platform, as seen in systems such as Halide~\cite{ragan-kelley2017}, TVM~\cite{chen2018d}, and RISE/ELEVATE~\cite{hagedorn2020}.
The schedule is expressed in a domain-specific language, where optimization choices such as decisions about the intermediate storage and the order of computation are defined by the programmer as transformations to code describing the algorithm.
The separation between high-level algorithm and platform specific transformations can allow more clear reasoning about the performance impact of optimizations when compared to defining the algorithm and its optimizations in the same code, as seen in system programming languages such as C, C++, and Rust.
Additionally, a single algorithm can have multiple schedules for different cases, such as hardware architectures or kernel properties, as seen in TVM which defines different schedules for the same algorithm for CPUs, GPUs, and specialized hardware accelerators~\cite{moreauHardwareSoftwareBlueprintFlexible2019,stjerngrenBifrostEndtoEndEvaluation2022}.

\begin{figure}[!t]
 \centering
 \includegraphics[width=0.65\columnwidth]{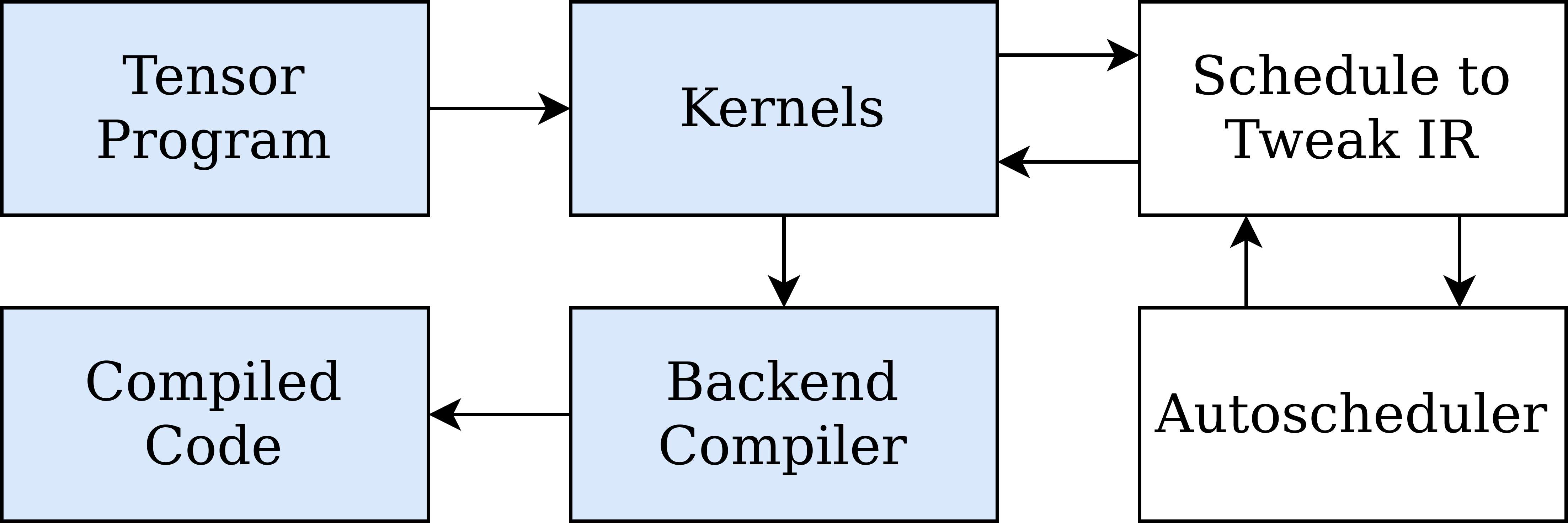}
 \caption{\label{fig:basic_schedule} Basic overview of the compute schedule paradigm for a tensor program. Non-shaded boxes are optional, but can improve performance significantly.}
\end{figure}

Figure~\ref{fig:basic_schedule} shows a tensor program being decomposed into kernels.
For example a kernel could be a layer of a neural network.
Each kernel corresponds to an \emph{algorithm} (a high level description of the desired computation), and can have a \emph{schedule} (the transformations to be applied to the algorithm) applied to it at compile time to produce efficient code on a target hardware platform.
These optimized kernels are composed together into a full program, where optimizations can include tiling, loop reordering, and where to apply vectorization.
The purpose of splitting a tensor program into kernels is so that they can be independently optimized.
This is valuable, since it makes the optimization problem more tractable, and kernels are assumed to be independent of each other.

TVM applies the compute schedule paradigm to the domain of DNN inference, exploiting the fact that the shapes of arrays are known beforehand, and thus integrating this knowledge when compiling models ahead-of-time.
% In TVM, DNN models can be loaded from a range of deep learning frameworks, including PyTorch~\cite{paszke2017automatic}, TensorFlow~\cite{abadi2016}, and ONNX~\cite{bai2019}.
TVM then applies pre-defined high-level optimizations to the whole computation graph, such as operator fusion (e.g., batch normalization layers can be completely removed from the graph for inference by combining their parameters with prior layers), and leverages its library of algorithms and \emph{schedules} to compile for backends including LLVM~\cite{llvm2004}, OpenCL~\cite{openCL}, CUDA~\cite{cuda2008}, and more.

However, writing optimized schedules by hand requires domain expertise, including understanding of the algorithm's design~\cite{gibson2020} and the behavior of the target hardware~\cite{zheng2018}.
In addition, an efficient schedule may not be efficient for all variants of the target algorithm, for example if the schedule was optimized assuming that the size of the inner loop would always be small.
In this case, optimizations which were used to exploit this assumed kernel characteristic might not help, and may even hinder performance.

Ansor~\cite{zheng2020} expands on the schedule model by introducing \emph{auto-schedules}, where efficient schedules for given kernels are generated automatically for a given hardware platform.
This reduces the need for specialized workload and platform expertise to produce efficient code, especially when new operations are introduced.
In this case, only the high-level algorithm needs to be written, and an efficient schedule for the target platform can be found via evolutionary search.
For this search, a space of potential schedules is explored (without needing to be defined by the designer, such as in AutoTVM~\cite{autotvm}) applying rules and random perturbations to find an efficient schedule for a particular instantiation of a given operation.
Auto-schedules in Ansor are tuned for individual kernels of a tensor program, with kernels being given a unique workload ID based on the hash of its key parameters (e.g., operation type, input data sizes).
If another tensor program contains an identical kernel (such a convolutional layer featuring the input and output dimensions), then the schedule can be reused, since the IDs of the kernels will match even if their weight data differs.
However, if these parameters change then a new tuning must be performed, since the kernel defines a different computation, thus having a different ID and optimal schedule.

%*********************************************************************************************

\section{Motivation}
\label{subsec:motivation}

\begin{figure}[t]
    \begin{subfigure}{\columnwidth}
     \centering
    \includegraphics[width=0.99\linewidth]{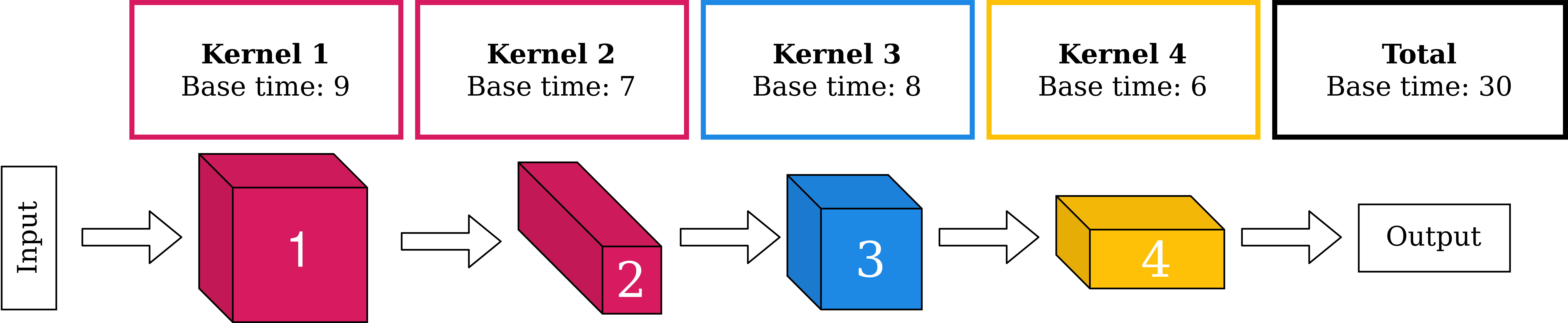}
    \caption{Kernels of a tensor program, 4 kernels of 3 classes.} \label{fig:tt_color_workloads}
      \end{subfigure}
 \par\bigskip % maximise vertical space here
 \begin{subfigure}{\columnwidth}
     \centering
    \includegraphics[width=0.99\linewidth]{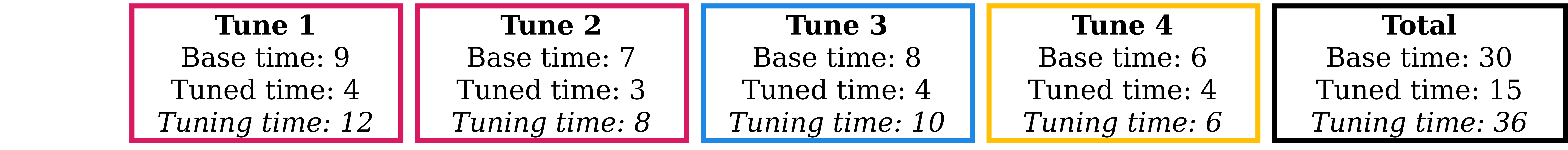}
    %\caption{Inference time and tuning costs when every workload fully tuned.}
    \caption{Inference time and tuning costs when kernels are fully tuned.}
    \label{fig:tt_full_tuning}
  \end{subfigure}
   \par\bigskip % maximise vertical space here
    \begin{subfigure}{\columnwidth}
     \centering
    \includegraphics[width=0.99\linewidth]{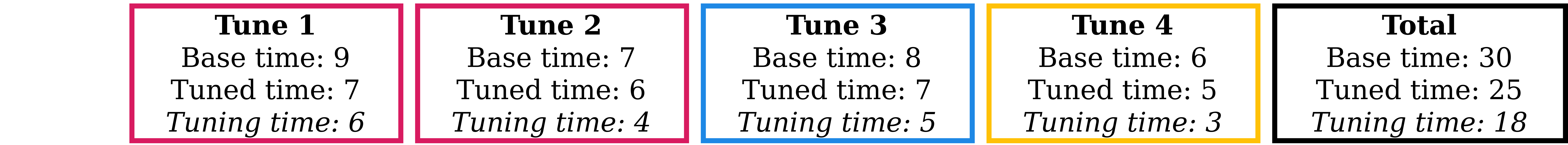}
    \caption{Inference time and tuning costs when tuning with reduced time.} \label{fig:tt_reduced_time}
  \end{subfigure}
   \par\bigskip % maximise vertical space here
    \begin{subfigure}{\columnwidth}
     \centering
    \includegraphics[width=0.99\linewidth]{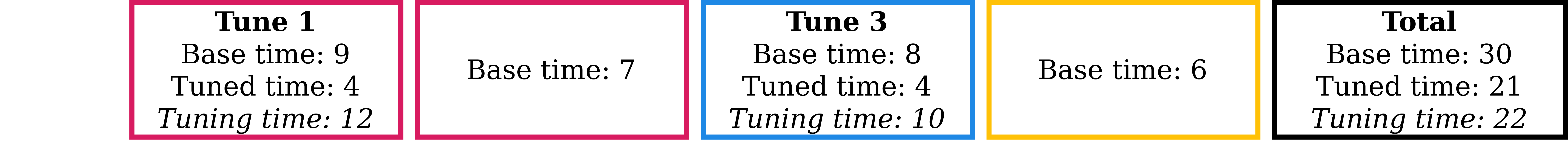}
    \caption{Inference time and tuning costs when tuning fewer kernels.} \label{fig:tt_partial_wkls}
  \end{subfigure}
    \par\bigskip % maximise vertical space here
    \begin{subfigure}{\columnwidth}
     \centering
    \includegraphics[width=0.99\linewidth]{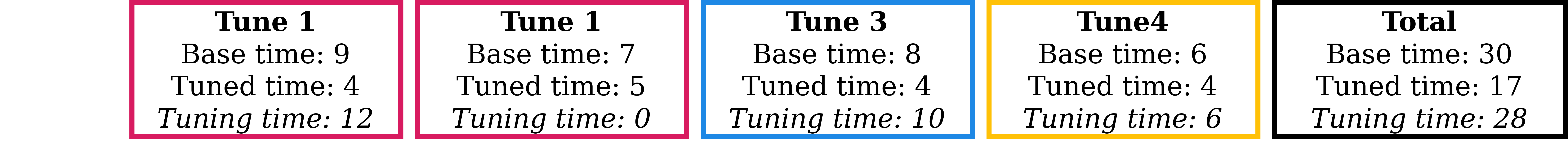}
    \caption{Inference time and tuning costs when using transfer-tuning to reuse the auto-schedule of Kernel 1 with Kernel 2.}
    %\caption{Inference time and tuning costs when tuning a subset of workloads, using transfer-tuning to reuse the auto-schedule of workload 1 with workload 2.}
    \label{fig:tt_transfer_tuning}
  \end{subfigure}
     \par\bigskip % maximise vertical space here
    \caption{Illustrative example of the costs and benefits of different approaches of auto-scheduling for a tensor program.}
  	\label{fig:colored_dag}%transfer_tuning_intiution}
\end{figure}

Auto-schedules, such as those provided by Ansor~\cite{zheng2020}, can yield state-of-the-art performance on a number of tensor programs (such as DNNs) and hardware devices by automatically generating optimized schedules.
Figure~\ref{fig:tt_color_workloads} shows an example of a simple computation Directed Acyclic Graph (DAG) for a tensor program, such as a DNN, with default untuned kernels.
Each node of the graph represents a unit of computation, also known as a \emph{kernel}.
In the figure colors represent the \emph{kernel classes}, meaning the types of operations contained within (e.g., DNN layer types such as convolutional layers), with there being three classes of kernels in this example and four kernels in total.
The shapes of the nodes represent the size of the data computed in the kernel, which in this example varies between kernels.
In this illustrative example if we auto-schedule this model, as shown in Figure~\ref{fig:tt_full_tuning}, we see that we can speed up the inference time by $2\times$ with a tuning time of $36$, where all values are for illustrative purposes.
%, and tuning time is several orders of magnitude higher than inference time.

However, it should be noted that auto-scheduling can be a very time consuming process, as shown in Figure~\ref{fig:motivation_time}, where the tuning time can be on the order of several hours for a whole DNN model depending on the complexity of the tensor program, the number of schedule variants chosen to evaluate, and the resources available of the target hardware device.
This high cost can be a bottleneck to deployment, since if we want to get the best inference time for a given DNN model we must spend a long period of time exploring schedule space.
Approaches to reduce search time include: reducing the time we allow the tuner as shown in Figure~\ref{fig:tt_reduced_time}; tuning a subset of the model's kernels as shown in Figure~\ref{fig:tt_partial_wkls}; or some combination of the two.
These trade-offs allow users to sacrifice potential improvements in performance for reduced tuning time.

Our main observation in Figure~\ref{fig:colored_dag} is that for kernels of the same class (where class is represented as color), since they define the same high level algorithm over varying data sizes, they may produce auto-schedules with similar properties.
% At a high level the algorithm, the only thing that will vary is the number of iterations, and the values of the elements of the tensors, which are generally not relevant to the high level description of the algorithm~\footnote{Sparse computations can add some caveats to this assumption.}.
A question posed by our observation is \emph{``Could an auto-schedule from a given kernel be reused on a different kernel of the same class?''}
% The question posed by our observation is \emph{``what would be the performance impact of using a schedule from a different workload of the same class?''}
If so, a further question is \emph{``How different will the optimizations found via auto-scheduling be between two kernels of the same class but varying data sizes?''}
The answer to the second question will vary depending on the structure of the computations defining the class, with factors such as access patterns and costs of the loop body playing important roles as well as the architecture of the target platform that the auto-schedule exploits, since the organization of the memory hierarchy and features such as vector-instruction size may make some optimizations more or less relevant.
Perhaps having some dimensions being similar (such as the extent of the innermost loop) could be more important than others (such as the extent of the outermost loop).

% Therefore, we define the process of reusing an auto-schedule for a given kernel on a different kernel as \emph{transfer-tuning}.
Therefore, answering the first question, we define the process of reusing an auto-schedule for a given kernel on a different kernel as \emph{transfer-tuning}.
In Figure~\ref{fig:tt_transfer_tuning} we show an example of transfer-tuning, where we reduce tuning time by reusing auto-schedules.
We reuse the auto-schedule for Kernel $1$ with Kernel $2$ to reduce the inference time without requiring any additional tuning.
We also use the auto-schedule of Kernel $1$ with itself, which we refer to as the ``native schedule''.
Note that we should expect some penalty when running Schedule $1$ with Kernel $2$ compared to running a natively auto-schedule for Kernel $2$, since a native schedule will exploit the specific data sizes of the computation to find optimizations for the target hardware and data size specific optimizations.
This kernel specific information would not be exploited by using Schedule $1$ for Kernel $2$, as the schedule is tuned for Kernel $1$.
The target of transfer-tuning is to improve the inference time of the overall tensor program, while being cheaper than running an auto-scheduler.
The trade-off between search time and performance improvement is interesting to explore and exploit, as long search times may not always be acceptable.
For example, a developer of AI applications for smartphones may not have the resources to provide auto-scheduling for their DNN model for the wide range of heterogeneous devices their app will be deployed on.
Nor will it be likely that smartphone users be willing to wait several hours for the DNN model to auto-schedule on their device.
However, in this case transfer-tuning could provide some performance speedups in a shorter period of time.
This reduced search time may also translate to reduced energy usage, as auto-scheduling is an energy intensive process that can saturate all CPU cores.
However in this work we focus purely on the inference time performance improvements of transfer-tuning.

% perry

% In Section~\ref{sec:contr} we discuss the concept of transfer-tuning in more detail, describing what is necessary for it to work, and how it might be implemented.

%*********************************************************************************************

\section{Transfer-Tuning}
\label{sec:contr}

%**********************************************************************************************

First, in Section~\ref{subsec:understanding_tt} we discuss some of the types of optimizations used by tensor program auto-schedulers, how transfer-tuning is possible, and how it can be beneficial.
Then in Section~\ref{subsec:workload_classes}, we further discuss the idea of kernel classes introduced in Section~\ref{subsec:motivation} and how they are a key part of transfer-tuning.
In Section~\ref{subsec:approach} we explore some of the behaviors of transfer-tuning on a full DNN model (ResNet18), and in Section~\ref{subsec:table_selection} we discuss some other practical considerations for transfer-tuning.

%**********************************************************************************************

\subsection{Principles of Transfer-Tuning}
\label{subsec:understanding_tt}

To understand how transfer-tuning works, we must first briefly explain the relevant concepts of schedules and auto-schedules. Let us consider an operation such as a matrix-multiply, which has a fixed loop structure but may have varying data sizes, as shown in lines $1-5$ of Algorithm~\ref{alg:gemm-sched}.

There are a variety of code transformations which can be applied to this operation, some of which are applicable to all instantiations of the operation and others which are specific to a particular input matrix size. 
For example, some transformations are valid regardless of the data sizes involved, such as if we instruct a given loop to be unrolled to its maximum depth.
No matter how many iterations are in a loop, it can be applied as long as we know the number of iterations ahead of time, thus it is valid regardless of if there are $3$ iterations or $300,000$.
However, the performance benefit of the transformation will be different depending on the number of iterations, with relevant factors including the architecture of the underlying hardware (e.g., the features of its cache) and properties of the computation such as the cost of the loop body.

In contrast, some transformations may only be valid for a given data shape.
For example, if we have a loop over the range $(0, N)$ where $N=32$, we could apply a loop splitting optimization defined as $\texttt{Split}(N, 4, 8)$ that breaks the loop into two loops in the ranges $(0, 4)$ and $(0, 8)$, which would allow us to traverse the full $32$ elements.
If we try to apply this optimization to a similar loop where $N=128$, then splitting it into the two prior ranges would produce invalid code, since we will not be able to cover the full space of the loop.
However, if we reformulate our transformation such that we apply it as $\texttt{Split}(N, (N/8), 8)$ our transformation becomes valid for all programs where $\{N \in \mathbb{N} : 8\mid N\}$.

Therefore some transformations can be applied to a kernel regardless of the data-shape, others can be reformulated to be valid for more than one data-shape, and some may not be valid for any data-shape other than the one they were originally defined for.
The performance benefits of these transformations may be data-shape dependent, for example a loop unrolling that brings benefit for a small loop range could bring a penalty for a larger loop range.
However, we claim that even with large data-shape differences some of these transformations can potentially improve performance, as compared to a generic schedule.

As discussed in Section~\ref{sec:background}, the process of auto-scheduling takes a set of kernels representing operations in a tensor program (such as a DNN) and iteratively explores the space of transformations that we can apply to each of them.
In this paper, when we apply the schedule produced for a given kernel via auto-scheduling and apply it to a kernel other than the one the schedule was tuned for, and we call this technique \emph{transfer-tuning}.
Transfer-tuning exploits the fact that many schedule transformations can be formulated to be data-shape agnostic, meaning that we can adapt schedules for kernels that they were not tuned for.

Let us take a look at a simple example of an auto-scheduled kernel, a row-major square matrix-multiply as defined in Algorithm~\ref{alg:gemm-sched}.
For more complex kernels (such as kernels containing convolutional layers) auto-schedules can be verbose, difficult to interpret, and intuitions as to why they provide good performance may be unclear, since they are automatically generated to exploit hardware performance dynamics that may not be evident, such as cache behavior.
We look at two data sizes for the operation, $C_1=A_1B_2$ which multiplies two $(512\times512)$ matrices, and $C_2=A_2B_2$ which multiplies two $1024\times1024$ matrices.
We use Ansor to produce auto-schedules for the two kernels, observing an improvement of $246\times$ and $308\times$ for $C_1=A_1B_2$ and $C_2=A_2B_2$ respectively compared to using an unoptimized schedule.
The auto-scheduling of the two kernels produces different schedules since they have different sizes.
Additionally, auto-scheduling in Ansor is non-deterministic due to the use of genetic algorithms to mutate schedules, and a stochastic learned cost model to reduce evaluation costs; thus differences in the auto-schedule may emerge even when re-running Ansor for the same kernel.
Lines~\ref{alg:first-step2}-\ref{alg:last-step2} and \ref{alg:first-step3}-\ref{alg:last-step3} of Algorithm~\ref{alg:gemm-sched} show a simplified representation of auto-schedules generated for $C_1=A_1B_1$ and $C_2=A_2B_2$ respectively on an Intel Xeon E5-2620 CPU.
Next, we briefly explain the schedule primitives used in this example, which is a subset of all the primitives available to Ansor and by extension, transfer-tuning.
%and transfer-tuning.

\begin{algorithm}[ht] %[htbp]
\footnotesize
  \caption{Auto-schedules for a GEMM operation \label{alg:gemm-sched}} 
  \begin{algorithmic}
    \State $\bm{A}:$ input matrix of size $N \times K$
    \State $\bm{B}:$ input matrix of size $K \times M$ 
    \State $\bm{C}:$ output matrix of size $N \times M$
  \end{algorithmic}
    \textbf{Unmodified row-major matrix-multiply computation}
  \begin{algorithmic}[1]
    \For{$n \gets 0$ to $N$}
        \For{$m \gets 0$ to $M$}
            \State $C[n][m] \gets 0$   \Comment{Initialize output value to zero}
            \For{$k \gets 0$ to $K$}
                \State $\bm{C}[n][m] \mathrel{+}= \bm{A}[n][k] \times \bm{B}[k][m]$
            \EndFor
        \EndFor
    \EndFor \label{alg:last-step1}
  \end{algorithmic}

  \textbf{Simplified auto-schedule where $N=P=K=512$}
  \begin{algorithmic}[1]
    \setcounterref{ALG@line}{alg:last-step1}
    \State $N_o, N_i \gets \texttt{Split}(N, 8)$ \label{alg:first-step2} %\Comment{$N_i$ now over range of $8$}
    \State $N_{oo}, N_o \gets \texttt{Split}(N_o, 1)$   \Comment{note $N_o$ redefined}
    \State $N_{ooo}, N_{oo} \gets \texttt{Split}(N_{oo}, 16)$ 
    
    \State $M_o, M_i \gets \texttt{Split}(M, 8)$
    \State $M_{oo}, M_o \gets \texttt{Split}(M_o, 1)$
    \State $M_{ooo}, M_{oo} \gets \texttt{Split}(M_{oo}, 16)$
    
    \State $K_o, K_i \gets \texttt{Split}(K, 1)$
    
    \State $\texttt{Reorder}(N_{ooo}, M_{ooo}, N_{oo}, M_{oo}, K_o, N_o, M_o, K_i, N_i, M_i)$
    
    \State $F_{NM} \gets \texttt{Fuse}(N_{ooo}, M_{ooo})$
    \State $\texttt{Parallel}(F_{NM})$
    \State $\texttt{Unroll}(F_{NM}, 512)$ \label{lst:line:unroll_512}
    
    \State $\texttt{Vectorize}(M_i)$ \label{alg:last-step2} % \Comment{$M_i$ is inner most loop, row-major data-layouts of $B$ and $C$ mean vectorisation makes sense}
  \end{algorithmic}
  \label{alg:gemm_512_schedule}
  
  \textbf{Simplified auto-schedule where $N=P=K=1024$}
  \begin{algorithmic}[1]
  \setcounterref{ALG@line}{alg:last-step2}
    \State $N_o, N_i \gets \texttt{Split}(N, 32)$  \label{alg:first-step3}
    \State $M_o, M_i \gets \texttt{Split}(M, 256)$
    \State $\texttt{Reorder}(N_o, M_o, N_i, M_i)$
    \State $\hat{N} \gets N_i, \hat{M} \gets M_i$
    \State Create Local Cache Buffer $\bm{D}$ of size $\hat{N}\times\hat{M}$ \label{lst:line:cache_buff}

    \State $\hat{N}_o, \hat{N}_i \gets \texttt{Split}(\hat{N}, 1)$
    \State $\hat{N}_{oo}, \hat{N}_o \gets \texttt{Split}(\hat{N}_o, 16)$
    \State $\hat{N}_{ooo}, \hat{N}_{oo} \gets \texttt{Split}(\hat{N}_{oo}, 2)$
    
    \State $\hat{M}_o, \hat{M}_i \gets \texttt{Split}(\hat{M}, 8)$
    \State $\hat{M}_{oo}, \hat{M}_o \gets \texttt{Split}(\hat{M}_o, 4)$
    \State $\hat{M}_{ooo}, \hat{M}_{oo} \gets \texttt{Split}(\hat{M}_{oo}, 8)$
    
    \State $K_o, K_i = \texttt{Split}(K, 4)$
    
    \State $\texttt{Reorder}(\hat{N}_{ooo}, \hat{M}_{ooo}, \hat{N}_{oo}, \hat{M}_{oo}, K_o, \hat{N}_o, \hat{M}_o, K_i, \hat{N}_i, \hat{M}_i)$
    
    \State $\texttt{ComputeAt}(\bm{D}, M_o)$
    \State $F_{NM} \gets \texttt{Fuse}(N_o, M_o)$
    \State $\texttt{Parallel}(F_{NM})$
    \State $\texttt{Unroll}(F_{NM}, 64)$ \label{lst:line:unroll_1024}
    \State $\texttt{Vectorize}(\hat{M}_i)$  \label{alg:last-step3}
  \end{algorithmic}
  \label{alg:gemm_1024_schedule}
\end{algorithm}
\begin{algorithm}[htbp]
\footnotesize
\caption{High level definition of a kernel class with a convolutional layer, bias addition, and ReLU activation.}\label{alg:wkl_class}
  \begin{algorithmic}[1]
\State Define placeholders for inputs $X$, weights $W$, and bias $B$
\State Pad the input $X\prime \gets$ \texttt{Pad}($X$)
\State $Y \gets$ \texttt{Conv2d}($X\prime$)
\State $Y \gets Y$ + $B$
\State $Y \gets$ \texttt{ReLU}($Y$)
\State Return $Y$
\end{algorithmic}
\end{algorithm}
\begin{itemize}[topsep=2.2pt]
    \setlength\itemsep{2.2pt}
    \item $\texttt{Split}(\text{[range]}, \text{[factor]})$: split a loop range into inner and outer ranges.
    %, with the inner range being the factor, and the outer range being the quotient.

    \item $\texttt{Reorder}(\text{[set of ranges]})$: specify a reordering of a set of nested loops.

    \item $\texttt{Fuse}(\text{[range]}, \text{[range]})$: fuse two consecutive loop ranges into a single range.

    \item $\texttt{Parallel}(\text{[range]})$: mark an axis to be used for multi-th\-r\-e\-a\-d\-ed computation.

    \item $\texttt{Unroll}(\text{[range]}, \text{[max unroll factor]})$: unroll a loop range up to a maximum depth.
\end{itemize}
\begin{itemize}[topsep=0pt]
    \item $\texttt{Vectorize}(\text{[range]})$: apply SIMD vectorization to a loop range.
    %, only valid if loop variable traverses contiguous data.

    \item $\texttt{ComputeAt}(\text{[output tensor]}, \text{[axis]})$: move a loop body computation such that it is computed at a given axis.
    %, used for fusing subsequent computations (e.g. moving activation functions to the body of the previous computation), or specifying the behavior of cache buffers.
\end{itemize}

Applying transfer-tuning to these GEMM computations, i.e. using the schedule generated for $C_1=A_1B_1$ with $C_2=A_2B_2$ and vice-versa, we observe that we still produce valid code, obtain performance within $5\%$ of the native tuning for both kernels, and a speedup of nearly $270\times$ when compared to the unmodified computation without a schedule.
The core difference in the auto-schedules produced for $C_1=A_1B_1$ and $C_2=A_2B_2$ is that the latter uses a temporary cache buffer to store intermediate results, as seen on Line~\ref{lst:line:cache_buff}.
Other differences are the unroll factors chosen by the auto-scheduler, $512$ for $A_1B_1$ as seen in Line~\ref{lst:line:unroll_512} and $64$ for $A_2B_2$ in Line~\ref{lst:line:unroll_1024}.
Note that in this case, when applying transfer-tuning all of the transformations being applied are still valid, since both computations are defined with the same initial loop structure and no transformation is strongly dependent on a given data size.

% Additionally, when we apply the schedule, we see a performance difference of $<5\%$ when compared to the native tuning, and a speedup of near $270\times$ when compared to the unmodified computation without a schedule.

% Transfer-tuning is a technique that re-uses the auto-schedule for a similar workload to improve the inference time for a workload without tuning that workload.
% This process is likely to produce less performance improvement than tuning the workload natively.
% It also has the possibility of increasing the inference time of the workload, if the schedule is not appropriate, for example when a loop unrolling transformation causes a large penalty.
% In this case, the auto-schedule should not be applied, and a fallback schedule should be used.

%**********************************************************************************************

\subsection{Kernel Classes}
\label{subsec:workload_classes}

We briefly introduced the idea of kernel classes in Figure~\ref{fig:colored_dag}, where we can reuse auto-schedules between kernels if they contain the same operations.
We now discuss the concept in more detail.
Kernels are the units of computation which we pass to the auto-scheduler, for example in DNNs a kernel may be a layer, or a set of layers that can be composed together.
Often kernels can contain several operations, especially when they can be fully fused to encompass the same loop structure (such as in the case of many activation functions like ReLU).
In this paper, we implement transfer-tuning using TVM, and defer to the kernel partitioning generated by TVM for a given DNN model, since the choices it makes are reasonable (such as combining activation functions and bias additions with larger layers such convolutional layers) and leads to state-of-the-art performance in many benchmarks~\cite{chen2018d}.

For example, a convolutional layer followed by a ReLU activation function can be treated as a single kernel, since we can fuse the operations such that the ReLU function is applied within the loop nest of the convolutional layer as soon as all the partial sums for an output have been computed.
This operation fusion saves a full traversal of the output data and can thus reduce cache misses significantly.
The purpose of having distinct kernels, rather than treating the whole program as a single function to be optimized, is that it allows the kernels to be optimized independently and in parallel.
We define a \emph{kernel class} to be a set of kernels that share the same sequence of operations, regardless of their data sizes.
For example, one kernel class could be characterized by containing only convolutional layers, another by containing a composition of dense and ReLU layers, etc.

\begin{table}[t]%[htbp]
    \centering
    \fontsize{6}{9}\selectfont
    \caption{Features of kernels in ResNet18, where \emph{class} is a label for the operations in the kernel (seen in \emph{TVM Ops}). 
    %\emph{Use Count} represents how many instances of said workload there are in the model. 
    \label{tab:resnet18_workload_features}}
% \begin{tabular}{cccccc}
% \begin{tabular}{|c|c|c|c|c|c|}
% \begin{tabular}{|Q{0.03\linewidth}|Q{0.07\linewidth}|Q{0.16\linewidth}|Q{0.16\linewidth}|Q{0.23\linewidth}|Q{0.07\linewidth}|}
\begin{tabular}{|M{0.03\linewidth}|M{0.06\linewidth}|M{0.16\linewidth}|M{0.18\linewidth}|M{0.22\linewidth}|M{0.07\linewidth}|}
\hline
 \textbf{ID} & \textbf{Class} & \textbf{input\_shape} & \textbf{kernel\_shape} & \textbf{TVM Ops} & \textbf{Use Count}  \\
\hline
\textbf{1} & A & $[1, 256, 14, 14]$ & $[512, 256, 7, 7]$ & conv2d\_add & 1\\ \hline
\textbf{2} & A & $[1, 128, 28, 28]$ & $[256, 128, 14, 14]$ & conv2d\_add & 1\\ \hline
\textbf{3} & A & $[1, 64, 56, 56]$ & $[128, 64, 28, 28]$ & conv2d\_add & 1\\ \hline
\textbf{4} & E & $[1, 3, 224, 224]$ & $[64, 3, 112, 112]$ & conv2d\_bias\_relu & 1\\ \hline
\textbf{6} & E & $[1, 64, 56, 56]$ & $[64, 64, 56, 56]$ & conv2d\_bias\_relu & 2\\ \hline
\textbf{7} & F & $[1, 64, 56, 56]$ & $[64, 64, 56, 56]$ & conv2d\_bias\_add\_relu & 2\\ \hline
\textbf{8} & E & $[1, 64, 56, 56]$ & $[128, 64, 28, 28]$ & conv2d\_bias\_relu & 1\\ \hline
\textbf{9} & E & $[1, 128, 28, 28]$ & $[128, 128, 28, 28]$ & conv2d\_bias\_relu & 1\\ \hline
\textbf{10} & F & $[1, 128, 28, 28]$ & $[128, 128, 28, 28]$ & conv2d\_bias\_add\_relu & 2\\ \hline
\textbf{11} & E & $[1, 128, 28, 28]$ & $[256, 128, 14, 14]$ & conv2d\_bias\_relu & 1\\ \hline
\textbf{12} & E & $[1, 256, 14, 14]$ & $[256, 256, 14, 14]$ & conv2d\_bias\_relu & 1\\ \hline
\textbf{13} & F & $[1, 256, 14, 14]$ & $[256, 256, 14, 14]$ & conv2d\_bias\_add\_relu & 2\\ \hline
\textbf{14} & E & $[1, 256, 14, 14]$ & $[512, 256, 7, 7]$ & conv2d\_bias\_relu & 1\\ \hline
\textbf{15} & E & $[1, 512, 7, 7]$ & $[512, 512, 7, 7]$ & conv2d\_bias\_relu & 1\\ \hline
\textbf{16} & F & $[1, 512, 7, 7]$ & $[512, 512, 7, 7]$ & conv2d\_bias\_add\_relu & 2\\ \hline
\end{tabular}

\centering
%\begin{tabularx}{\linewidth}{|C|C|C|C|C|C|}
% \begin{tabular}{|c|c|c|c|c|c|}
% \begin{tabular}{|Q{0.03\linewidth}|Q{0.07\linewidth}|Q{0.16\linewidth}|Q{0.16\linewidth}|Q{0.23\linewidth}|Q{0.07\linewidth}|}
\begin{tabular}{|M{0.03\linewidth}|M{0.06\linewidth}|M{0.16\linewidth}|M{0.18\linewidth}|M{0.22\linewidth}|M{0.07\linewidth}|}
\hline
 \textbf{ID} & \textbf{Class} & \textbf{input\_shape} & \textbf{pool\_size} & \textbf{TVM Ops} & \textbf{Use Count} \\
\hline
\textbf{5} & B & $[1, 64, 112, 112]$ & $[2, 2]$ & max\_pool2d & 1\\ \hline
\textbf{17} & C & $[1, 512, 7, 7]$ & $[7, 7]$ & global\_avg\_pool2d & 1\\ \hline
\end{tabular}
\\
% \begin{tabular}{|c|c|c|c|c|c|}
% \begin{tabular}{|Q{0.03\linewidth}|Q{0.07\linewidth}|Q{0.16\linewidth}|Q{0.16\linewidth}|Q{0.23\linewidth}|Q{0.07\linewidth}|}
\begin{tabular}{|M{0.03\linewidth}|M{0.06\linewidth}|M{0.16\linewidth}|M{0.18\linewidth}|M{0.22\linewidth}|M{0.07\linewidth}|}
 \hline
 \textbf{ID} & \textbf{Class} & \textbf{input\_shape} & \textbf{weights\_shape} & \textbf{TVM Ops} & \textbf{Use Count} \\ \hline
\textbf{18} & D & $[1, 512]$ & $[1, 1000]$ & dense\_add & 1\\
\hline
\end{tabular}

\end{table}

In Table~\ref{tab:resnet18_workload_features} we observe the characteristics of the kernels and their classes in the ResNet18~\cite{he2016deep} model, a Convolutional Neural Network (CNN) defined on the ImageNet dataset~\cite{ILSVRC15}.
Most kernels include a 2D convolutional layer, some of which include an activation function, or a bias or a skip-connection addition.
However we also observe some pooling layers and a fully-connected layer.
Some kernels are repeated more than once in the model\footnote{Note that in ResNet18 the $18$ refers to the total number of convolutional and fully connected layers.}, as represented by the ``Use Count'' column.
However, for the purposes of auto-scheduling repeated kernels are only tuned once, although may be given a higher proportion of the search time.
Overall in ResNet18 we identify $6$ kernel classes, labeled A-F: with classes A, E, and F representing kernels featuring convolutional layers; B and C being max-pooling and average-pooling layer kernels; and D being the final fully-connected layer.
We highlight that there are a variety of kernel classes featuring convolutional layers (class labels A , E, and F), with kernels of class E also including a bias addition, followed by a ReLU activation, an overview of which we describe in Algorithm~\ref{alg:wkl_class}.
The operations of class E can be further decomposed into lower level loop structures such as those describing the \texttt{Conv2d} algorithm, however we do not include these details for brevity.

Along with characterizing operations of its class, kernels are also defined by the data size of their inputs and weights.
A schedule for a kernel would apply transformations to the code in a manner similar to the one seen in Algorithm~\ref{alg:gemm-sched}, albeit with the transformations being applied to a more complex initial loop structure.
Much like the GEMM example in Section~\ref{subsec:understanding_tt}, we observe that schedules can be reused between kernels of the same class in ResNet18, even if they are defined with different sizes.
Thus, we could run transfer-tuning using a schedule of class E on another kernel of class E.
In some cases the generated code may be invalid, for example if the schedule defines a loop splitting factor which is larger than the loop itself.
Attempting to apply a schedule from class E to another class, such as one defined by a fully-connected (dense) layer of class D, would always be invalid as the schedule would try to apply transformations to computations and loops not present in the computation.
In principle, for kernel classes which share some of the operations (e.g., classes E and F), their schedules could be adapted to allow a form of across-class transfer-tuning.
Exploration of this idea, as well as its impact on performance are outside the scope of this paper.

%In Section~\ref{subsec:across_class_tt} we discuss how a weak form of across-class transfer-tuning may be possible, however for now we assume transfer-tuning is only valid between workloads of the same class.

% \begin{figure}[t]
%   \centering
% \begin{subfigure}{0.31\columnwidth}
%   \centering
%   \includegraphics[width=\columnwidth]{fig/transfer_tuning_overview-ansor.png}
%   \caption{Ansor}
%   \label{fig:workflow_ansor}
%  \end{subfigure}
%  \hspace{1em}%
% \begin{subfigure}{0.31\columnwidth}
%   \centering
%   \includegraphics[width=\columnwidth]{fig/transfer_tuning_overview-us.png}
%   \caption{Transfer-tuning}
%   \label{fig:workflow_us}
% \end{subfigure}
% %\caption{Put your caption here}
% \caption{Auto-scheduling workflow in Ansor, compared to transfer-tuning.  We reduce tuning time by exploiting similarities in the data shapes to tune a smaller number of workloads.}
% \label{fig:workflow}
% \end{figure}

%**********************************************************************************************

\subsection{Applying transfer-tuning}
\label{subsec:approach}

Now that we have discussed the principles of transfer-tuning, including how it works and why kernel classes are relevant, next we look at how it performs using a real DNN model.
We take the ImageNet definition of ResNet18 and use the auto-schedules of ResNet50 generated by Ansor, a model chosen because it is likely to have good potential for transfer-tuning due to its similar structure.

% Our end-to-end transfer-tuning implementation has two stages: first we evaluate each workload/schedule pair as standalone programs; second we use this information to test if the best workload/schedule pairs chosen behave differently in the context of the full model.

%----------------------------------------------------------------------------------------------

% \subsubsection{Standalone execution}

\begin{figure*}[t]
 \centering
 \includegraphics[width=0.98\textwidth]{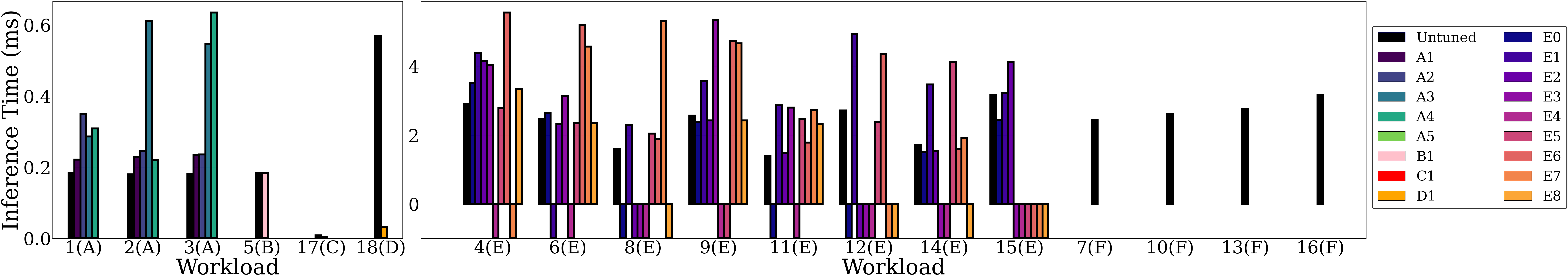}
 \caption{\label{fig:standalone_wkls} Inference time of ResNet18 kernels using ResNet50 schedules, with lower being better. Negative values denote schedules which produced invalid code.}
\end{figure*}

First, we evaluate each of the $18$ kernels of ResNet18 with all compatible schedules of ResNet50.
Figure~\ref{fig:standalone_wkls} shows the inference time of all of these kernel/schedule pairs, running as distinct programs.
The purpose of this evaluation is to give us insights into which schedules provide good performance improvements for each kernel.
We also compare against the performance of the kernel when it uses the default schedule provided by TVM, which we refer to as ``untuned''.

We observe that there are six kernel classes in ResNet18, with no schedules for classes F found in ResNet50.
For class F we use the default schedule provided by TVM, represented as a black bar.
For kernels of class A we have $4$ compatible schedules to try from ResNet50, kernels of class E can be compiled with $16$ possible schedules, and for kernels of classes B, C, and D note that we only have one compatible schedule each.
For class E we observe that some schedules on some kernels produce invalid code, which we represent with a value of $-1$.
There are $16$ schedules of class E from which $7$ always produce invalid code for the kernels of ResNet18, hence we do not represent them in the graph.
We also observe significant differences in inference time between schedules for some kernels, for example for kernel $2$ schedule A3 has over double the inference time of A4.

% Therefore, running standalone is a reasonable proxy for performance in the full model, since the executed code of workloads in this context are independent function calls.
% Their only point of interaction is the output data of one workload being used as input data of a subsequent workload, which may have consequences for data locality.
% The independence of workloads assumption is used by Ansor, which tunes all workloads in isolation and takes the best performing schedule for each.
% This is pragmatic, as it means that many more schedules can be evaluated due to reduced search costs, increasing the likelihood that a performant one can be found.
% However, our observation is that the independence assumption is too strong.

Taking the best schedule found via transfer-tuning for each kernel of ResNet18 and using them when compiling the full model we can observe a speedup of $1.2\times$, as shown in the leftmost bar of Figure~\ref{fig:speedup}.
The bar next to it shows that given the same search time Ansor can only achieve a speedup of $1.01\times$.
Search time for transfer-tuning means the time for testing each kernel of the target model with each valid schedule of the model chosen for transfer-tuning, and choosing the best.
This search time (around $1.2$ minutes for ResNet18) is shown in Figure~\ref{fig:search_time}.
In addition, we compare how long Ansor requires to achieve our speedup, which in the case of ResNet18 is $4.8\times$ longer.
This validates that transfer-tuning can work in the context of a full DNN model.
However, we chose the model to tune with (ResNet50) arbitrarily.
Thus in Section~\ref{subsec:table_selection} we give an overview of how we might select a model in a more systematic manner.

%**********************************************************************************************

\subsection{Model selection}
\label{subsec:table_selection}

In Section~\ref{subsec:approach} we demonstrated the core concepts of transfer-tuning using the ResNet18 model tuned from ResNet50.
This was a reasonable choice, since the model architectures are similar (they belong to the same family of models).
However, we need a more robust approach to select the model we will use for transfer-tuning, which we explore using 10 other models.

%---------------

\subsubsection{Selection heuristic}
\label{subsec:selection_heuristic}

Table~\ref{tab:model_features} shows a set of DNN models, their kernel classes, the frequency of kernels of each class, and the proportion of the untuned inference time that kernels of a given class represent.
For example, ResNet50 has $6$ kernel classes representing $27$ unique kernels (some kernels are repeated in the model), with kernels of class E representing the majority of the untuned inference time ($67\%$), and kernels of classes B, C, and D representing a negligible proportion of the inference time.
%, with classes including: class A with four workloads, which represents $17\%$ of the untuned inference time; $16$ workloads of class E representing $67\%$ of the inference time; and classes B, C, and D which represents a negligible proportion of the inference time.
For brevity we do not include details of each class, however expensive kernel classes tend to include convolutional layers or fully connected layers, while cheaper classes tend to contain operations such as pooling layers.

When choosing a model we want to maximize the likelihood that it will provide a good tuning for a target model.
We hypothesized in Section~\ref{subsec:motivation} that perhaps similarities between the kernels (e.g., having the same convolutional kernel size, or similar memory footprint) could be used to predict how successful a given transfer-tuning for a kernel using a given schedule would be.
However, in our initial study we did not find any feature which had strong predictive power.
Thus, in this paper we adopt a more coarse-grained approach which chooses a model to tune from based on the number of available schedules of a given class, and the proportional cost of that kernel class in untuned inference in the target model.

We define a selection heuristic which for a target model chooses a model to tune with that maximizes the number of available tuned schedules, giving preference for kernel classes that represent a higher proportion of the untuned inference time.
To avoid models with very high numbers of schedules dominating the heuristic, we increase the influence of the untuned costs by squaring it and reduce the influence of the number of schedules in the tuning model by taking the square root.
Thus, we formulate our heuristic for a given target model $M$, which has a set of kernel classes $C$, as choosing a tuning model $T$ which maximizes the following:

\begin{equation} \label{eqn:heur}
\sum_{c \in C}^{} P_c^2 \sqrt{|W_{Tc}|},
\end{equation}

where $P_c$ is the proportional cost of kernel class $c$ in $M$, and $W_{Tc}$ is the set of kernels of class $c$ in the candidate model $T$.
Looking at Table~\ref{tab:model_features}, we can see for ResNet50 that the model which maximizes Equation~\ref{eqn:heur} is GoogLeNet, and the two versions of EfficientNet maximize each other.
For BERT and MobileBERT it is clear why they are chosen for each other, as both contain kernels of class Q (containing only a ``dense'' operation) representing $98\%$ and $97\%$ of the inference time respectively.

However this heuristic is not guaranteed to make optimal decisions.
For example, the heuristic chose GoogLeNet for ResNet50 in part because it had a high number of schedules for class E, which represents $67\%$ of ResNet50's untuned inference time.
This means that for each of the $16$ kernels of class E in ResNet50 there are $49$ schedules that may reduce the inference time.
However, the $9$ schedules of class E in VGG-16 may be better at reducing the overall inference time in ResNet50, even though there are fewer of them. 
Note that the heuristic could be improved if we had a better predictive model of which schedules may perform well for transfer-tuning, however in this work we observe that this basic heuristic gets reasonable results.
% Section~\ref{sec:discussion} discusses how a more nuanced model selection could be made in the future, including mixing workloads from more than one model.

Table~\ref{tab:heur_choices} shows transfer-tuning's maximum speedup by applying the top $3$ models suggested by the heuristic.
As we can see, the trend is that the best speedup is achieved by Choice $1$, and the maximum speedup decreases with subsequent options.
Note that for BERT and MobileBERT every other model ties for second and third place and gives no speedup, hence we leave these entries blank (represented with ``-'').
This is because the only kernel class in BERT and MobileBERT shared by other models is class D, which represents less than $0.1\%$ of their inference time.

\begin{table*}[ht]
\footnotesize
\centering
\caption{Kernel classes of DNN models, with the number of kernels of each class, and the proportion of the untuned inference time these kernels represent.
Also shown is the model chosen for transfer-tuning.
\label{tab:model_features}}
\begin{tabular}{|c|l|l|l|}
\hline
  \textbf{ID} & \textbf{Model} & \textbf{Kernel classes (number of kernels, percentage of inference time)} & \textbf{Tuning Model} \\
\hline
\textbf{M1} & ResNet50 & \textbf{A} (4, 17\%); \textbf{B} (1, 0\%); \textbf{C} (1, 0\%); \textbf{D} (1, 6\%); \textbf{E} (16, 67\%); \textbf{G} (4, 10\%) & GoogLeNet \\ \hline
\textbf{M2} & AlexNet & \textbf{B} (3, 0\%); \textbf{D} (1, 6\%); \textbf{E} (5, 14\%); \textbf{H} (2, 80\%); \textbf{I} (1, 0\%) & VGG-16 \\ \hline
\textbf{M3} & VGG-16 & \textbf{B} (5, 0\%); \textbf{D} (1, 1\%); \textbf{E} (9, 59\%); \textbf{H} (2, 40\%); \textbf{I} (1, 0\%) & GoogLeNet \\ \hline
\textbf{M4} & MobileNetV2 & \textbf{A} (7, 15\%); \textbf{C} (1, 0\%); \textbf{D} (1, 24\%); \textbf{J} (8, 32\%); \textbf{K} (5, 15\%); \textbf{L} (10, 14\%) & EfficientNetB4 \\ \hline
\textbf{M5} & EfficientNetB0 & \textbf{A} (14, 9\%); \textbf{C} (11, 4\%); \textbf{D} (1, 12\%); \textbf{K} (5, 9\%); \textbf{M} (8, 39\%); \textbf{N} (12, 27\%); \textbf{O} (7, 0\%) & EfficientNetB4 \\ \hline
\textbf{M6} & EfficientNetB4 & \textbf{A} (16, 11\%); \textbf{C} (13, 3\%); \textbf{D} (1, 10\%); \textbf{K} (7, 14\%); \textbf{M} (9, 39\%); \textbf{N} (14, 23\%); \textbf{O} (9, 0\%) & EfficientNetB0 \\ \hline
\textbf{M7} & GoogLeNet & \textbf{B} (10, 1\%); \textbf{C} (1, 0\%); \textbf{D} (1, 4\%); \textbf{E} (49, 95\%) & ResNet50 \\ \hline
\textbf{M8} & MnasNet1.0 & \textbf{A} (7, 17\%); \textbf{D} (1, 25\%); \textbf{E} (9, 31\%); \textbf{K} (5, 15\%); \textbf{P} (12, 13\%) & GoogLeNet \\ \hline
\textbf{M9} & BERT & \textbf{D} (1, 0\%); \textbf{Q} (3, 98\%); \textbf{R} (2, 2\%); \textbf{S} (1, 0\%); \textbf{T} (1, 0\%); \textbf{U} (1, 0\%); \textbf{V} (1, 0\%)  & MobileBERT \\ \hline
\textbf{M10} & MobileBERT & \textbf{D} (1, 0\%); \textbf{Q} (4, 97\%); \textbf{R} (2, 3\%); \textbf{S} (1, 0\%) & BERT \\ \hline
\end{tabular}

\end{table*}

\begin{table}[ht]
\begin{threeparttable}[ht]
\footnotesize
\centering
\caption{Transfer-tuning performance in terms of speedup using the top 3 choices from the heuristic.}\label{tab:heur_choices}
\begin{tabularx}{\columnwidth}{|X|C|C|C|}%|c|c|c|c|}
\hline
% \textbf{Model}          & \multicolumn{2}{c}{\textbf{Choice 1} &\multicolumn{2}{c}{\textbf{Choice 2}} & \multicolumn{2}{\textbf{Choice 3}}   \\
% \hline
% ResNet50      & \textbf{M7} & \textit{$1.17\times$} & \textbf{M8} & \textit{$1.0\times$} & \textbf{M3} & \textit{$1.09\times$} \\

\textbf{Model}          & \textbf{Choice 1} & \textbf{Choice 2} & \textbf{Choice 3}   \\
\hline
ResNet50      & \textbf{M7} (\textit{$1.16\times$}) & \textbf{M8} (\textit{$1.0\times$}) & \textbf{M3} (\textit{$1.09\times$}) \\ \hline

 AlexNet       & \textbf{M3}  (\textit{$4.6\times$}) & \textbf{M7} (\textit{$1.05\times$}) & \textbf{M1} (\textit{$1.03\times$})  \\ \hline
 VGG-16         & \textbf{M7} (\textit{$1.19\times$}) & \textbf{M1} (\textit{$1.0\times$}) & \textbf{M8} (\textit{$1.0\times$}) \\ \hline
 MobileNetV2   & \textbf{M6} (\textit{$1.20\times$}) & \textbf{M5} (\textit{$1.21\times$}) & \textbf{M8} (\textit{$1.19\times$}) \\ \hline
 EfficientNetB0 & \textbf{M6} (\textit{$1.23\times$}) & \textbf{M4} (\textit{$1.08\times$}) & \textbf{M8} (\textit{$1.09\times$}) \\ \hline 
 EfficientNetB4 & \textbf{M5} (\textit{$1.13\times$}) & \textbf{M4} (\textit{$1.04\times$})  & \textbf{M8} (\textit{$1.03\times$}) \\ \hline
 GoogLeNet     & \textbf{M1} (\textit{$1.15\times$}) & \textbf{M3} (\textit{$1.04\times$}) & \textbf{M8} (\textit{$1.0\times$}) \\ \hline
MnasNet1.0    & \textbf{M7} (\textit{$1.26\times$}) & \textbf{M1} (\textit{$1.25\times$}) & \textbf{M3} (\textit{$1.18\times$}) \\ \hline
BERT     & \textbf{M10} (\textit{$59\times$}) & -\tnote{$\dagger$} & -\tnote{$\dagger$} \\ \hline
MobileBERT    & \textbf{M9} (\textit{$13\times$}) & -\tnote{$\dagger$}  & -\tnote{$\dagger$}  \\
\hline
\end{tabularx}

\begin{tablenotes}
%\item[$\dagger$] Note that models M1-M8 tie according to the heuristic, and all get a speedup of $1.0\times$
\item[$\dagger$] Note that models M1-M8 tie for Choices 2 and 3 giving no speedup (i.e., $1.0\times$), hence we leave these entries blank (represented with “-”).
\end{tablenotes}

\end{threeparttable}
\end{table}

%---------------

\subsubsection{Alternative heuristics}
\label{subsec:improved_heuristics}

\begin{figure*}[t]
\centering
    \begin{subfigure}{0.49\textwidth}
        \centering
        \includegraphics[width=\textwidth]{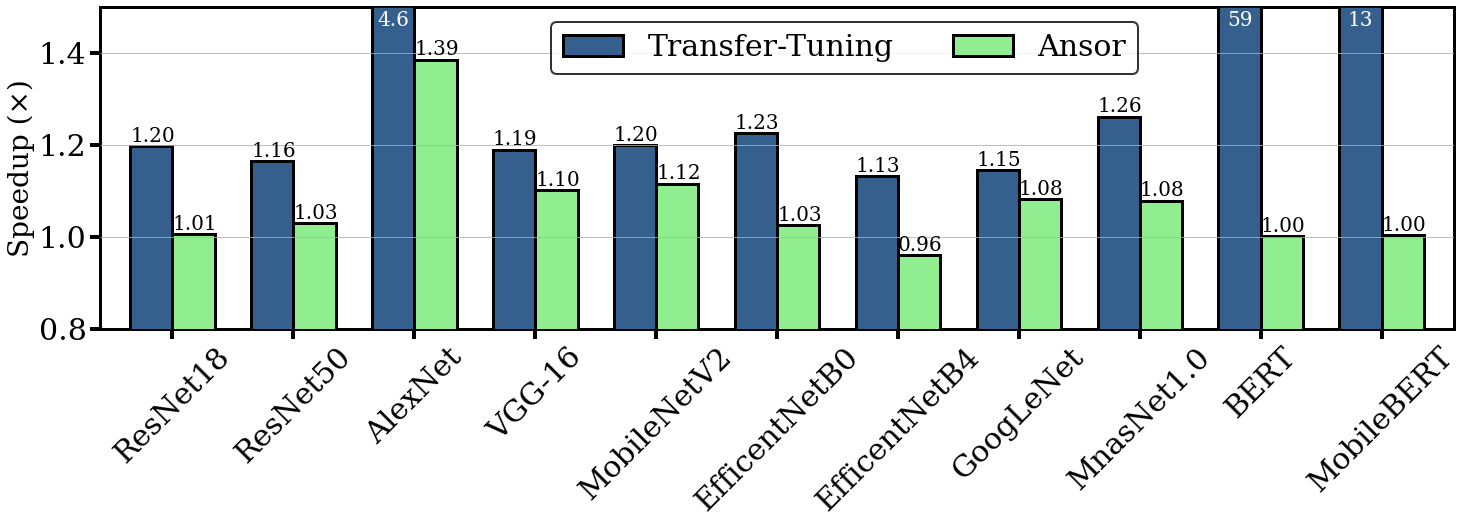}
        \caption{Speedup for transfer-tuning and Ansor given the same search time.} \label{fig:speedup}
  \end{subfigure}%
  \hspace*{\fill}   % maximize separation between the subfigures
  \centering
    \begin{subfigure}{0.49\textwidth}
    \centering
        \includegraphics[width=\textwidth]{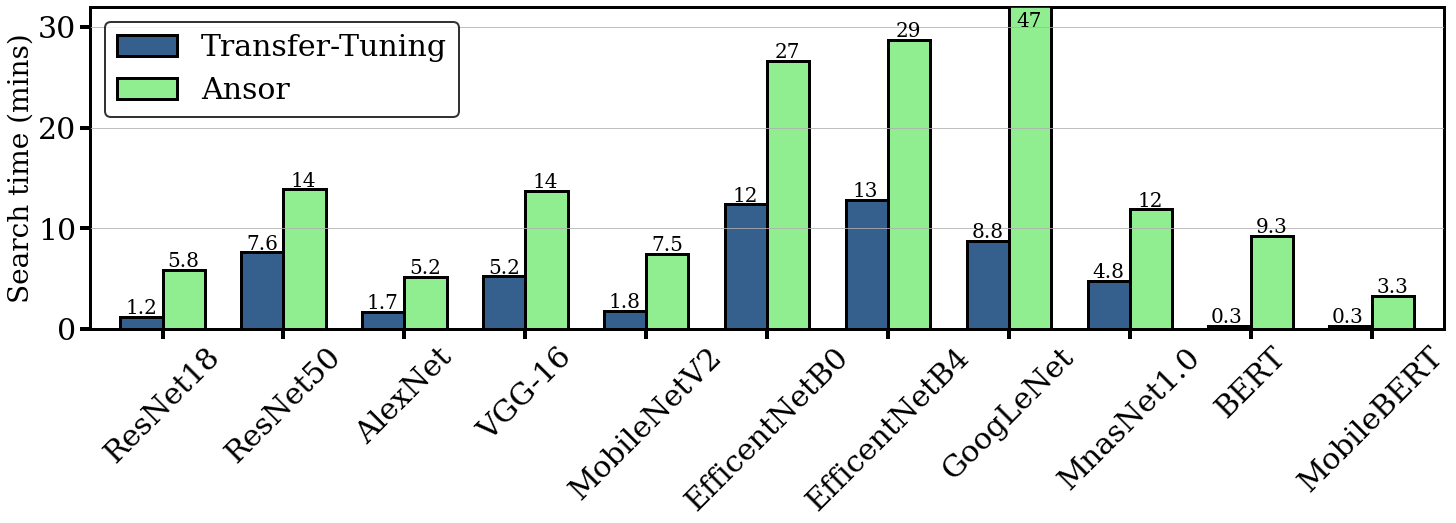}
        %\caption{Search time for transfer-tuning, and the search time required for Ansor to match our speedup.} \label{fig:search_time}
        \caption{Search time for transfer-tuning, and Ansor to match its speedup.} \label{fig:search_time}
    \end{subfigure}%
    \caption{Transfer-tuning results for several models on a server-class CPU (Intel Xeon E5-2620).}
  	\label{fig:oto_models-pact}
\end{figure*}

In Figure~\ref{fig:oto_models-pact} we provide an evaluation of the choices made by the heuristic described in Equation~\ref{eqn:heur}, demonstrating that it can outperform tuning the DNN models from scratch with Ansor.
However, we could explore extensions to this heuristic which may allow greater exploitation of transfer-tuning, improving the speedup and/or reducing the search time.
For example, the heuristic chooses a single model to transfer-tune from, however in principle we could use all of the tuned schedules we have available in Table~\ref{tab:model_features}.
We evaluate the impact of using all the available tuned schedules in Section~\ref{subsec:robustness}.
The caveat to consider is that this could translate into a very high number of schedules to evaluate, which would increase search time significantly.
Thus, a more intelligent heuristic might discard schedules that are less likely to improve performance, and prioritize kernel classes by the potential improvement they could get, since we observe that the average speedups achievable by different kernel classes vary.
We will explore this, and other potential extensions to transfer-tuning in future work.

% Additionally, we could leverage information about the differences in potential speedup for each workload class

%*********************************************************************************************

\section{Evaluation}
\label{sec:evaluation}

% After introducing the core ideas of transfer-tuning, we now look at how the process can be applied to other DNN models.

In this section we evaluate the performance of $11$ common DNN models, ResNet18 and $10$ more shown in Table~\ref{tab:model_features}, applying transfer-tuning using auto-schedules from the model selected using the heuristic described in Section~\ref{subsec:table_selection}.
In Sections~\ref{subsec:rpi4}, \ref{subsec:bert_seq_len}, and \ref{subsec:robustness} we explore transfer-tuning on an edge platform, varying the sequence length, and the impact of using a larger pool of schedules.

%We analyze the impact of increasing the number of full model evaluations and compare how well Ansor auto-scheduling from scratch performs when given the same search time.

%**********************************************************************************************

\subsection{Experimental Setup}
\label{subsec:exp-setup}

In addition to ResNet18 discussed in Section~\ref{sec:contr} we evaluate $10$ more DNN models.
The first $8$ are CNNs defined on the ImageNet dataset~\cite{ILSVRC15} and the final $2$ being Transformer-based~\cite{vaswani2017attention} models for natural language sequence classification.
The machine used to evaluate the models includes an 8 core Intel Xeon E5-2620 CPU.
Auto-scheduling, compilation, and inference are executed using the CPU, using 1 thread per CPU core.
For our baselines, we take the median inference time for each model over 10 runs, compiled using TVM's standard untuned schedules and the \texttt{-o3} flag.

\textbf{ResNet18} and \textbf{ResNet50}~\cite{he2016deep} have $18$ and $50$ layers respectively and consist of residual blocks.
Each block contains two convolutional layers (that include between $64$ and $2048$ filters of size $3\times3$ and $1\times1$) and blocks are connected in a feed-forward manner.
% Additionally, there is a \textit{skip connection} between the input and output of each block, which helps gradients propagate through the network~\cite{shatteredgradients2017}.
% These models also utilize batch normalization~\cite{ioffe2015batch} after each convolution, though these layers are combined with the parameters of the convolutional layers such that they have zero cost. %, and can be effectively removed from the graph.

\textbf{AlexNet}~\cite{krizhevsky2012} is a canonical CNN model and consists of $5$ convolutional, $3$ max-pooling, and $3$ fully connected layers.
Newer DNNs use fewer fully connected layers to increase efficiency.
% Each convolutional layer consists of convolutional filters (different sizes are used including $3\times3$, $5\times5$ and $11\times11$) and a nonlinear activation function ReLU.
% It also uses batch normalization and a softmax function, though again the former does not have impact on inference time performance.

\textbf{VGG-16}~\cite{Simonyan15} is a CNN with $13$ convolutional layers and $3$ fully connected layers.
Some versions include batch normalization layers, but in TVM these are removed/fused for inference.
% Each convolution uses a $3\times3$ kernel and there are max-pooling operations after layers~\{2, 4, 7, 10, 13\}.
%, assuming the inference framework applies the appropriate graph level optimization, which TVM does.

\textbf{MobileNetV2}~\cite{sandler2018mobilenetv2} is a lighter weight model with $53$ layers, many of which feature depthwise convolutions which reduce the number of parameters and operations required.
This makes it ideal for constrained edge devices.

\textbf{EfficientNet}~\cite{tan2019c} is a family of models with a focus on scalability.
%variants of which continue to rank highly in the state-of-the-art for a range of standard datasets.
The architecture of the smallest model (EfficentNetB0) was found using neural architecture search (NAS)~\cite{zoph2017}, and the accuracy of the model is improved by applying a novel scaling method which changes the architecture to efficiently increase the number of parameters and operations.
There are sizes ranging from B0-B7, and in this paper we evaluate EfficentNetB0 and EfficentNetB4.

\textbf{GoogLeNet}~\cite{szegedy2015} (or InceptionV1) is a $22$ layer model containing $9$ so-called ``inception'' modules.
This technique allows a deeper model to be trained more efficiently.
% It came first in the ILSVRC 2014 image classification challenge.

\textbf{MnasNet}~\cite{tan2019b} is an model architecture designed for edge devices such as mobile phones with an architecture generated using NAS.
We evaluate the model using a depth multiplier of $1.0$, which contains $52$ convolutional layers and a dense layer.

\textbf{BERT}~\cite{devlin2019a} is a Transformer-based~\cite{vaswani2017attention} model which excels in several natural language processing (NLP) tasks.
It contains $12$ layers, where a layer is a so called ``transformer block''.
We take a definition of BERT for sequence classification tasks.
%, however between tasks, in general this only changes the final operation.

\textbf{MobileBERT}~\cite{sun2020} is a compressed model inspired by the BERT architecture.
% It reduces the size by decomposing BERT into smaller building blocks and transposing the dimensions to produce a bottleneck effect to the original block.
It has $24$ layers and around $4.4\times$ fewer parameters than BERT.
For both BERT and MobileBERT which can take variable length input, we fix the sequence length at $256$, with a discussion of the impact of varying the sequence length in Section~\ref{subsec:bert_seq_len}.
% is a task-agnostic compressed version of \bert by building on the foundation of Knowledge Distillation. MobileBERT's main contribution is to separate BERT into smaller building blocks and transposing the dimensions to produce a bottleneck effect to the original block.

%**********************************************************************************************

\subsection{Results}
\label{subsec:across_model_tt}

%Figure~\ref{fig:oto_models} shows the results of running transfer-tuning across the $8$
Figure~\ref{fig:oto_models-pact} shows the results of running transfer-tuning across the $11$ models, with each model being tuned using the model suggested by our heuristic described in Section~\ref{subsec:table_selection}.
The only exception to this is ResNet18, which was used as in illustrative example in Section~\ref{sec:contr}.
Figure~\ref{fig:speedup} shows the speedup achieved by transfer-tuning and Ansor given the same search time.
Figure~\ref{fig:search_time} shows the search time required by transfer-tuning and how much time Ansor requires to match transfer-tuning's speedup.

For ResNet50 we observe a speedup of $1.16\times$, requiring $7.2$ minutes to achieve.
Given the same search time, Ansor gets a speedup of $1.03\times$ and requires $1.8\times$ as much search time to reach the same speedup.
%we have $27$ workloads in total, with some workloads repeated throughout the model.
For AlexNet we observe a speedup of $4.6\times$ which takes $1.7$ minutes to achieve.
The maximum search time for AlexNet is lower than ResNet50, as it has a smaller number of kernels.
Ansor given the same search time gets a speedup of $1.39\times$ and requires $3.1\times$ more search time to reach the same speedup.
For VGG-16, we observe a speedup of $1.19\times$ which takes $5.2$ minutes to achieve.
To achieve the same speedup Ansor requires $2.6\times$ as much time.
%we have $12$ workloads across $5$ classes, with the tuning provided by GoogLeNet.

% VGG16 has $18$ workloads across $5$ classes.
% The heuristic tells us that we should select GoogLeNet, as seen in Table~\ref{tab:heur_choices}.
% For VGG16, the maximum speedup of $1.19\times$, with Ansor getting a speedup of $1.1\times$ given the same search time of $5.2$ minutes.

MobileNetV2 (tuned using schedules from EfficientNetB4) obtains a maximum speedup of $1.2\times$, which takes $1.8$ minutes.
Ansor given the same time gets a speedup of $1.12\times$ and requires $4.2\times$ more search time to reach the same speedup. % as MobileNetV2.
We also observe that over half of its kernels (those of classes J and L), representing around $46\%$ of the untuned inference time, are not transfer-tuned by EfficientNetB4 since it does not contain them.
This suggests that there is further scope for improvement, for example tuning using schedules from a model which included those kernel classes could increase the maximum speedup obtained.

EfficientNetB0 (also tuned with EfficientNetB4) gets a maximum speedup of $1.23\times$, which takes $12$ minutes.
The search time is much higher than most other models, since there are $58$ kernels to evaluate with 764 unique kernel/schedule pairs.
Ansor given the same time gets a speedup of $1.03\times$ and requires $2.15\times$ as much search time to reach the same speedup.

For EfficientNetB4 (tuned with EfficientNetB0) the situation is similar with a high search time of $13$ minutes due to having $69$ kernels, or 775 kernel/schedule pairs to evaluate.
The speedup is $1.13\times$ which is lower than EfficientNetB0, with Ansor requiring $2.23\times$ more time to reach the same speedup.
We observe that given the same time as transfer-tuning Ansor sees a slowdown by $0.96\times$ compared to the baseline.
This is not unexpected, as due to their stochasticity, auto-schedulers can sometimes hurt performance initially even if they eventually converge on an improved schedule.

For GoogLeNet, we observe a speedup of $1.15\times$ which takes $8.8$ minutes to achieve.
Like the two EfficientNet models, a higher number of kernels ($61$) make the search time higher than other models.
Given the same time, Ansor achieves a speedup of $1.08\times$ and requires $5.3\times$ more time to reach the same speedup.
% To achieve the same speedup Ansor requires $2.6\times$ as much time.
% GoogLeNet features $61$ workloads, and reaches a maximum speedup of $1.15\times$.

As the final ImageNet model, MnasNet1.0 tuned with GoogLeNet achieves a maximum speedup of $1.26\times$, taking $4.8$ minutes.
Given the same time, Ansor takes $1.08\times$ and requires $2.5\times$ as much time to achieve the same speedup.

Finally, BERT and MobileBERT see the most dramatic performance improvements of $59\times$ and $13\times$ respectively.
In addition, they see the largest relative difference in search time required compared to Ansor, $33\times$ and $10\times$ respectively.

%

%Section~\ref{subsec:vgg16} explores this poor performance.

Overall, these results show that transfer-tuning can outperform Ansor when given a limited amount of search time.
Figure~\ref{fig:motivation_time} shows that each model varies in the potential maximum speedup it can achieve, where we take the maximum speedup to be achieved using Ansor's recommended $20,000$ schedule variants.
Therefore, to compare the performance of our DNN models fairly we show the proportion of this maximum speedup transfer-tuning achieves in Table~\ref{tab:max_value}.
On average, transfer-tuning achieves $49.12\%$ of Ansor's maximum speedup, with VGG-16 being the lowest with $17.69\%$ and BERT being the highest with $88.41\%$
Compared to the search time required by Ansor to explore $20,000$ schedule variants, transfer-tuning requires only $2.08\%$ of this time on average.
However, the values in Figure~\ref{fig:search_time} give a more informative comparison showing that to achieve the same speedup as transfer-tuning, Ansor requires over $6.5\times$ more time than transfer-tuning on average, with the lowest relative difference being for ResNet50 ($1.8\times)$, and the highest being for BERT ($33\times$).

\begin{table}
\small
\centering
\caption{Transfer-tuning versus $20,000$ Ansor iterations.}\label{tab:max_value}
\begin{tabular}{|c|c|c|}
\hline
\textbf{Model} & \textbf{Speedup (\%)} & \textbf{Search time (\%)} \\
\hline
ResNet18 & 49.20 & 0.48 \\ \hline
ResNet50 & 40.65 & 2.91 \\ \hline
AlexNet & 71.54 & 0.64 \\ \hline
VGG-16 & 17.69 & 1.41 \\ \hline
MobileNetV2 & 29.16 & 1.10 \\ \hline
EfficentNetB0 & 80.14 & 5.34 \\ \hline
EfficentNetB4 & 52.35 & 6.41 \\ \hline
MnasNet1.0 & 44.18 & 2.40 \\ \hline
GoogLeNet & 48.00 & 2.00 \\ \hline
BERT & 88.41 & 0.08 \\ \hline
MobileBERT & 18.96 & 0.10 \\ \hline
\textbf{Mean} & \textbf{49.12} & \textbf{2.08} \\
\hline
\end{tabular}

\end{table}

\subsection{Exploring a constrained edge platform}
\label{subsec:rpi4}

To further validate transfer-tuning we evaluate our models on a Raspberry Pi 4 B, a common low-power edge device with an Arm Cortex-A72 CPU.
Such devices represent another potential application of transfer-tuning, as they may not have the resources to undertake auto-scheduling themselves.
Ansor allows edge devices to be connected to a more powerful server which runs auto-scheduling over RPC.
However, this process can still be slow, may not always be available, and is not scalable to deployment across large heterogeneous fleets of devices.

\begin{figure*}[t]
\centering
    \begin{subfigure}{0.49\textwidth}
        \centering
        \includegraphics[width=\textwidth]{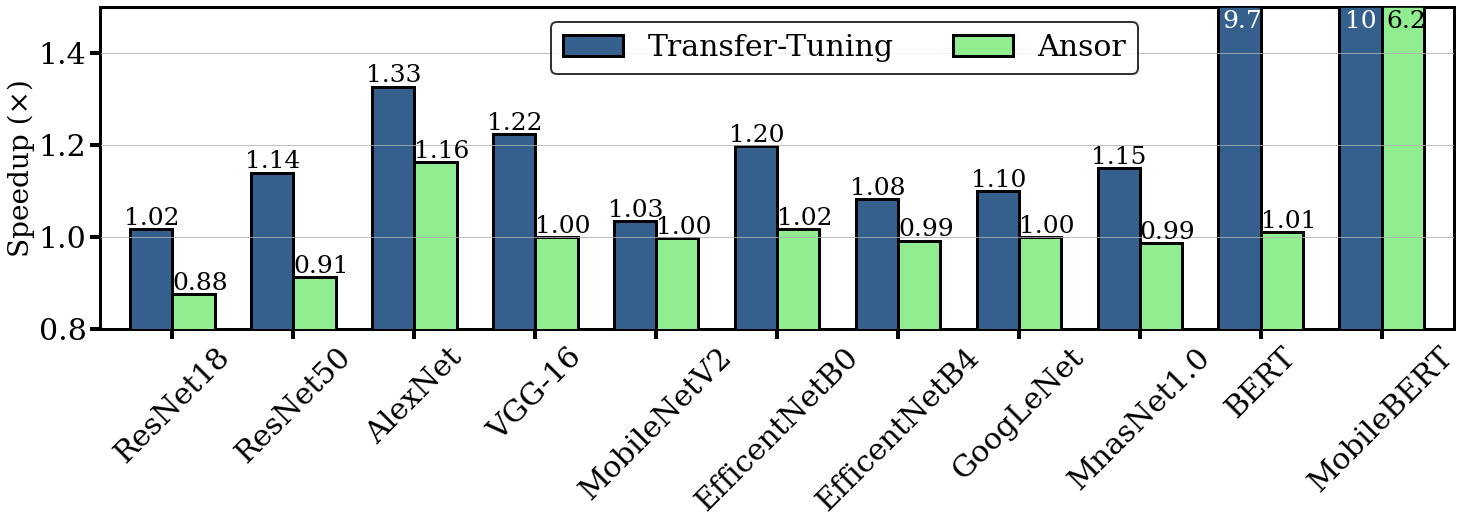}
        \caption{Speedup for transfer-tuning and Ansor given the same search time.} \label{fig:speedup_rpi4}
  \end{subfigure}%
  \hspace*{\fill}   % maximize separation between the subfigures
  \centering
    \begin{subfigure}{0.49\textwidth}
    \centering
        \includegraphics[width=\textwidth]{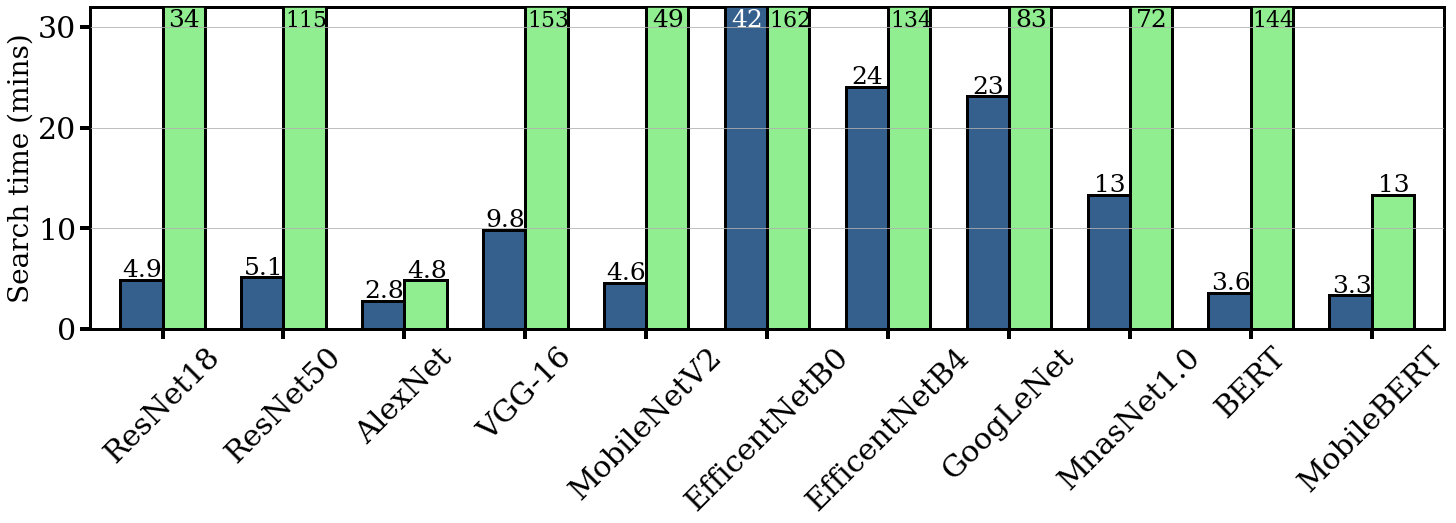}
        %\caption{Search time for transfer-tuning, and the search time required for Ansor to match our speedup.} 
        \caption{Search time for transfer-tuning, and Ansor to match its speedup.}
        \label{fig:search_time_rpi4}
    \end{subfigure}%
    \caption{Transfer-Tuning results for several models on an edge CPU (Arm Cortex-A72).}
  	\label{fig:models_rpi4}
\end{figure*}

Figure~\ref{fig:speedup_rpi4} shows the speedups achieved on the Raspberry Pi 4, and Figure~\ref{fig:search_time_rpi4} shows the search time required.
%We do not include EfficentNetB4, as we experienced errors running on this platform.
We observe that the relative differences between transfer-tuning and Ansor become exacerbated in terms of tuning time, with Ansor requiring over $10.8\times$ as much time to reach the same speedup on average, which is significantly higher than the $6.5\times$ difference observed on the x86 platform.
%Generally, Ansor sees negligible speedup, however AlexNet and MobileBERT see higher speedups.
In future work we will explore if transfer-tuning is viable between hardware platforms.

%Further study will tell us if this trend is consistent across constrained devices.
%, and if transfer-tuning is viable between varying CPU models.

%**********************************************************************************************

\subsection{Varying sequence length}
\label{subsec:bert_seq_len}

Unlike the ImageNet models, which takes input data of fixed sizes ($224\times224$), sequence models such as BERT and MobileBERT can take variable input sizes, e.g. a longer or shorter sentence.
However, from the perspective of Ansor varying the input size means the whole model is different, since every single kernel has different data sizes to process.
In principle, auto-scheduling could occur with a given dimension being specified as being dynamic.
However, to date TVM has poor support for this and no support for this when tuning~\footnote{See \url{https://discuss.tvm.apache.org/t/does-tvm-support-dynamic-input-shape/11069}}.
In addition, this could potentially lose out on some optimizations by keeping the input size fixed.

\begin{figure}[t]
    \centering
    \includegraphics[width=0.75\linewidth]{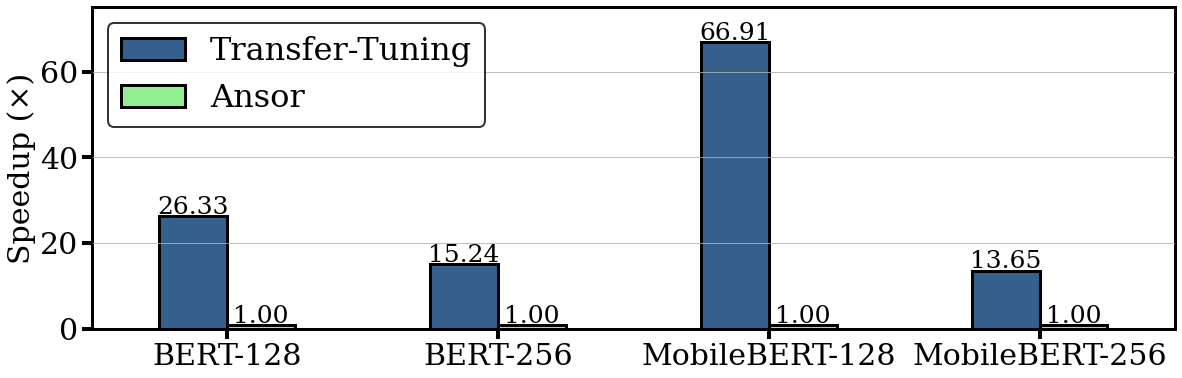}
    \caption{Transfer-tuning varying the sequence length of BERT models (Intel Xeon E5-2620).}
    \label{fig:tt_seq_len}
\end{figure}

Therefore, as an alternative view on transfer-tuning, we evaluate models of the same architecture but with different input sizes, namely BERT and MobileBERT for sequence lengths of $128$ and $256$.
In Section~\ref{subsec:across_model_tt} we evaluated these models for a sequence length of $256$, therefore we must also tune versions of these models with sequence length $128$.
Figure~\ref{fig:tt_seq_len} shows the results, with for example ``BERT-128'' representing the BERT model for sequence length $128$ being tuned using schedules from ``BERT-256''.

We observe that the improvement is greater applying tuning from a larger sequence length to a smaller sequence length, $3.3\times$ as much improvement on average.
We also note that compared to the results of Figure~\ref{fig:oto_models-pact} (which shows BERT and MobileBERT with sequence length $256$ being tuned with each other), BERT in Figure~\ref{fig:tt_seq_len} gets less of a speedup ($3.87\times$ less) and MobileBERT gets approximately the same speedup (around $13\times$).

Varying input data sizes are common in sequence models such as BERT and MobileBERT.
However, in CNNs for computer vision often publicly available models trained on a common dataset (e.g., ImageNet) are fine-tuned on a new dataset which may have a different input data size.
Thus, this could represent another use-case for transfer-tuning, which we leave for future work.

%**********************************************************************************************

\subsection{Alternative heuristics}
%\subsection{Alternative formulation}
\label{subsec:robustness}

As discussed in Section~\ref{subsec:improved_heuristics}, there is more than one way for transfer-tuning to select schedules, and this can be an implementation detail to suit the needs of a given use-case.
%transfer-tuning paradigm does not prescribe how tuned schedules should be selected, and thus can be an implementation detail to suit the needs of the use-case.
Throughout the paper we have demonstrated the core concepts and functionality of transfer-tuning by implementing ``one-to-one'' model transfer-tuning, described by our heuristic in Section~\ref{subsec:selection_heuristic}.
This heuristic was devised from analytical observations about the features of models and their kernel classes, and has demonstrated speedups successfully.
However, as discussed in Section~\ref{subsec:improved_heuristics}, if we have tuned schedules available for a set of DNN models, as an alternative approach we could explore exploiting all of these schedules regardless of model.

Thus in this section we provide a brief evaluation of how transfer-tuning can be implemented to deal with this possibility.
%, although we leave refinement and expansion of these methods for future work.
For each of our models described Section~\ref{subsec:exp-setup}, we take the pool of schedules from Table~\ref{tab:model_features} and make all of them available to the target model.
Note that the concept of ``models'' is irrelevant to the pool, and for every kernel in the target model transfer-tuning picks the best schedule according to the standalone performance of its kernel.

\begin{figure*}[ht]
\centering
    \begin{subfigure}{0.49\textwidth}
        \centering
        \includegraphics[width=\textwidth]{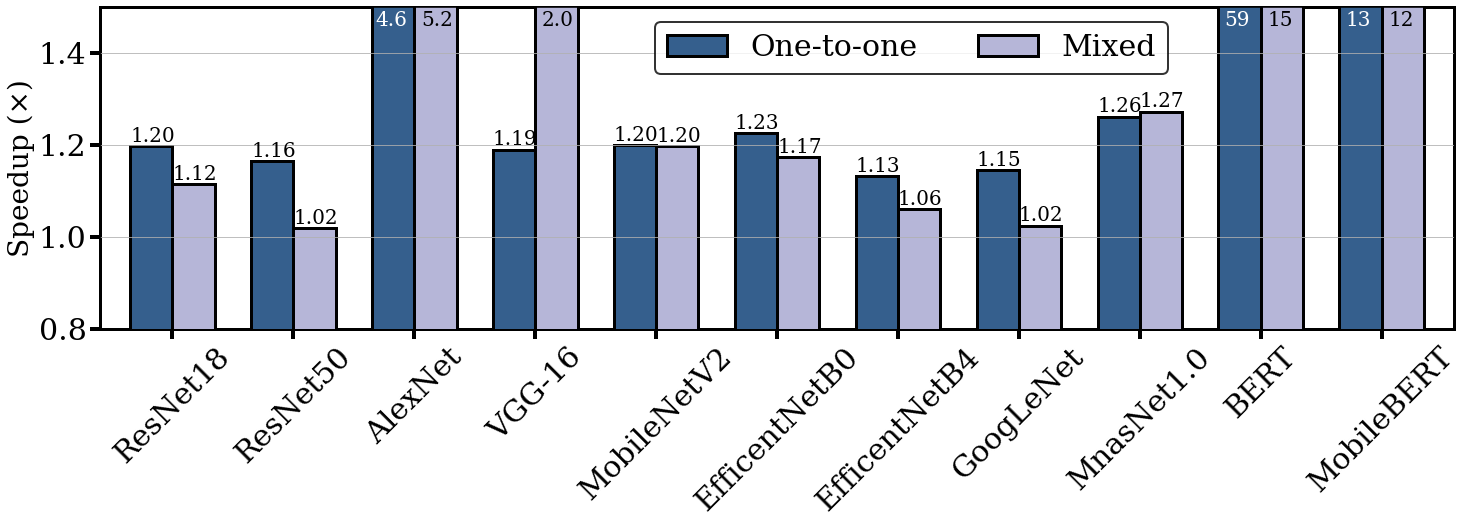}
        \caption{Speedup for transfer-tuning using schedules from a single model, and a mixed pool of models.} \label{fig:random_pool}
  \end{subfigure}%
  \hspace*{\fill}   % maximize separation between the subfigures
  \centering
    \begin{subfigure}{0.49\textwidth}
    \centering
        \includegraphics[width=\textwidth]{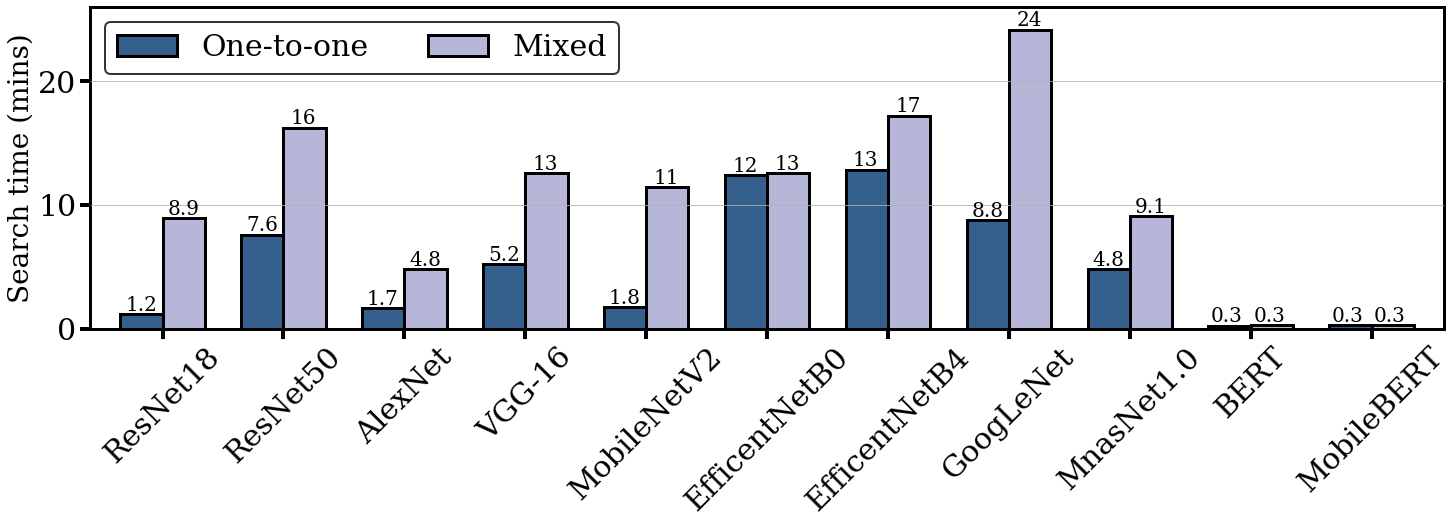}
        \caption{Search time for transfer-tuning using schedules from a single model, and a mixed pool of models.} \label{fig:pool_search_time}
    \end{subfigure}%
    \caption{Transfer-tuning using a schedule pool of several models on a server-class CPU (Intel Xeon E5-2620).}
  	\label{fig:pool-pact}
\end{figure*}

\vfill
We show the results of this evaluation in Figure~\ref{fig:pool-pact}.
Our first observation in Figure~\ref{fig:pool_search_time} is that the search time increases by around $2\times$ on average, with the highest being ResNet18 with a $5.34\times$ increase.
An increase in search time is expected, since we increased the number of kernel/schedule pairs we evaluate.
However, because each model contains varying kernel classes, this increase in pairs varies between models.
%the increase in the number of new kernel/schedule pairs varies between models.
For instance, BERT and MobileBERT see a negligible increase in search time, as only kernels of class D are given new schedules to explore.
In situations with many kernel/schedule pairs, we could reduce the search time by sampling a subset of schedules, either randomly or using some other selection heuristic.

\vfill
For speedup, we can make several observations from Figure~\ref{fig:random_pool}.
First we see that the maximum speedups achieved for AlexNet, VGG-16, and MnasNet1.0 increase compared to ``one-to-one''.
For MnasNet1.0 the improvement is modest ($1.27\times$ speedup compared to $1.26\times$), however for AlexNet and VGG-16 it is more significant: from $4.6\times$ to $5.2\times$ and from $1.19\times$ to $2.0\times$ respectively.
This makes intuitive sense: either we use the best schedules found in the one-to-one approach, or we have new schedules in the pool that allow us to improve further.
We observe that MobileNetV2 chooses the same schedules as before and provides the same speedup.

%achieves the same performance in this setting, and chooses the same schedules as before.

% \begin{figure*}[ht]
% \centering
% \includegraphics[width=\textwidth]{fig/pact/tt_random_pool.png}
% \caption{\color{red} Speedup for transfer-tuning all schedules from Table~\ref{tab:model_features}, as compared against using a single model (one-to-one method seen in Figure~\ref{fig:speedup}), running on a server-class CPU (Intel Xeon E5-2620).
% } \label{fig:random_pool}
% \end{figure*}

% \begin{figure}[t]
% \centering
% \includegraphics[width=\linewidth]{fig/pact/tt_full_pool.png}
% \caption{\color{red} Speedup for transfer-tuning all schedules from Table~\ref{tab:model_features}, as compared against using a single model (one-to-one method seen in Figure~\ref{fig:speedup}), running on a server-class CPU (Intel Xeon E5-2620).
% } \label{fig:random_pool}
% \end{figure}

\vfill
However, contrary to this intuition, we observe that $7$ models see a \emph{reduced} speedup.
% This is surprising: surely the best schedule/kernel pairs being selected can only match or improve our previous performance?
Despite the fact we are selecting the kernel/schedule pairs with the lowest standalone inference time, our overall speedup when running the full model is lower than the initial one-to-one approach, even though the kernel/schedules used in the one-to-one case gave higher standalone inference times.
%\vspace{1mm}
\vfill
Our conclusion is that although the performance of kernels running in a standalone manner are a proxy to their performance in the context of a full tensor program, they do not capture all potentially relevant interactions.
Our implementation of transfer-tuning assumes that the fastest kernel running as a standalone program will also be the fastest when running in the context of the full tensor program.
This assumption of the independence of kernels is also used by Ansor, which tunes all kernels as standalone programs, and combines the best schedules together.

It is clear that this assumption is sufficient for transfer-tuning to provide improvements over Ansor, however the results in Figure~\ref{fig:pool-pact} demonstrate that there may be performance considerations of ``inter-kernel'' relationships that are not captured by standalone kernel evaluation.
For example, the output data of one kernel may be used as the input data for a subsequent kernel.
The data access patterns of the first kernel will dictate the cache placement of the output data, which will impact the read times of the data when it is used in the second kernel.
We can imagine an extreme case where the average reuse distance of the output/input data between kernels is at its maximum.
This could significantly increase the inference time of the second kernel.

% Experimentally, transfer-tuning has demonstrated that it can function without this consideration.
% However stronger guarantees that a larger schedule pool size can only improve the performance is desirable, and awareness and exploitation of this dynamic may enable further optimizations.

Therefore, awareness and exploitation of this dynamic may enable further optimizations for transfer-tuning and related methods.
We leave a thorough exploration of inter-kernel relationships for future work, however approaches could include per-kernel profiling when running the full program, and evaluating kernels pairwise.
% If we observe a significant difference between a kernel's standalone inference time and its inference time in the full program, this tells us that there is likely such a dynamic occurring.
% Alternatively, we could evaluate kernels pairwise, and build up a set of schedules which minimizes the expected inference times.
% These approaches will need to balance the search costs this could incur against the potential benefits, and take opportunities to discard poorly performing schedules early in the search process.

%after discarding schedules which provide unreasonably high inference times in standalone.
%Alternatively, we can leverage kernel profiling when running the full model.

% that we can still achieve high speedups, even in this setting of lower-similarity transfer-tuning.
% This demonstrates that we can ``mix-and-match'' from a range of lower similarity models and the approach can still work.
% As discussed in Section~\ref{subsec:improved_heuristics}, improved ways of selecting tuned schedules from our pool could improve the quality of this approach even further.

%\color{black}

%*********************************************************************************************
% \input{06-discussion}
%*********************************************************************************************

\section{Related Work}
\label{sec:related_work}

% \subsection{Schedules, auto-tuning and auto-scheduling}

Schedule based computation was popularized in Halide~\cite{ragan-kelley2017}, however it does not provide all of the graph level optimizations available in TVM~\cite{chen2018d}.
TVM builds on the ideas of Halide to bring a usable schedule compiler for machine learning.
Other works include RISE/ELEVATE~\cite{hagedorn2020} which are well defined functional languages for compute declaration and scheduling respectively, however are not currently production ready.

Auto-tuning frameworks, especially for compute-schedule based systems like TVM, are a popular area of research.
%where specified template parameters are searched for a given program
AutoTVM~\cite{autotvm} takes hand-engineered schedules for operations and explores parameter tunings across a space defined by the schedule author, such as tiling sizes, unrolling factors, and others.
However, it should be noted that AutoTVM compared to Ansor cannot reach the same same maximum speedup, as AutoTVM constrains the search space.
Choices to efficiently explore AutoTVM's search space vary, including approaches such as gradient boosting~\cite{chen2016} and genetic algorithms.
Chameleon~\cite{ahn2020} improves the search strategies of AutoTVM by leveraging reinforcement learning.
In future work transfer-tuning could be extended to vary parameters from schedules transfer-tuned from another model to further increase performance.

% MetaTune~\cite{ryu2021} improves auto-tuning parameter selection for systems like AutoTVM. 
% In ATF~\cite{rasch2021}, the authors look at more efficient auto-tuning techniques with a focus on modeling when parameters have inter-dependencies among them.

% To adapt schedules to the diversity of workload sizes that a given algorithm may be parameterized by, AutoTVM~\cite{autotvm} improved the performance portability of their schedules by leveraging auto-tuning, by exploring the impact of varying schedule parameters such as tile sizes, or loop re-orderings on the target platform for a given tensor program.
% This can bring high performance improvements, however as well as being a time-consuming process, auto-tuning still requires the schedule to be hand-written.
% Additionally the schedule designer must define the search space to explore, which requires further expertise.
% The chosen space could be either too large to easily find an efficient set of tuned parameters, or too restrictive and leave out more optimal configurations. 
% For these reasons, auto-tuning hand-written schedules is difficult to scale, since schedules must be optimized for new hardware architectures, and novel algorithms must have schedules written from scratch, which requires large engineering efforts, which can be a barrier to the adoption of novel solutions such as novel DNN architectures~\cite{barham2019}.

%~\cite{lee2019a} leverages the internal program~\cite{lee2019a} leverages the internal program states to achieve better tuning.
 
Regarding auto-scheduling, FlexTensor~\cite{zheng2020a} is an auto-scheduling system similar to Ansor, although it relies on more hand written templates, thus seeing worse performance on some benchmarks.
LIFT~\cite{steuwer2015} explores the use of rewrite rules on high level representations of programs to generate OpenCL code, although it does not explore its large search space as efficiently as Ansor.
The Tiramisu deep learning compiler~\cite{baghdadi2019} recently added support for an auto-scheduling system~\cite{baghdadi2021}.
Overall, none of these approaches exploit the notion of transfer-tuning to reduce search time.

The reuse of bundles of optimizations has been explored in other works beyond tensor programs.
For example, Martins et al.~\cite{martins2016} looks at similarities between C functions to cluster them into groups and applies compiler passes based on group membership.
% They represent programs using the DNA symbolic representation~\cite{sanches2010}.
In transfer-tuning we have more domain specific knowledge we can leverage, since we are in the space of tensor programs with well-defined operations.
CompilerGym~\cite{CompilerGym} exposes LLVM compiler optimizations to reinforcement learning agents via the OpenAI Gym~\cite{brockman2016}. 
Like Martins et al.~\cite{martins2016}, its focus is on more general purpose program optimization and does not exploit domain specific knowledge of tensor programs.

Overall, transfer-tuning leverages the ideas of schedule based programming paradigms such as Halide and TVM, as well as auto-scheduling introduced by Ansor.
Other works have exploited similarity between programs to make compilation optimization more efficient.
However, transfer-tuning's novelty comes from leveraging this workload similarity in the domain of schedule based tensor compilers to reduce search costs.

\vfill

\section{Conclusion}
\label{sec:conclusion}

In this paper we proposed \emph{transfer-tuning} as a new approach to exploit similarities in tensor programs to reuse efficient schedules found via auto-scheduling. 
We have discussed how transfer-tuning is feasible in a compute/schedule programming paradigm, and explained the key components necessary to accelerate tuning of a full model.
We defined an implementation of transfer-tuning and evaluated the performance on $11$ models on a server-class x86 CPU, achieving $49.12\%$ of the maximum speedup achieved by the Ansor auto-scheduler on average, and with Ansor requiring over $6.5\times$ as much time to match our performance.
We also evaluated transfer-tuning on a constrained edge device, the Raspberry Pi 4, showing that the gap between Ansor and transfer-tuning is exacerbated, with Ansor requiring over $10.8\times$ as much time to match our performance.
%with a maximum performance speedup of between $1.16\times$ and $4.76\times$, while clearly outperforming Ansor when given limited search time.
% We also discussed how the simple implementation can be extended to better exploit the potential of transfer-tuning, such as incremental compilation, as well as some possible applications.
% For future work, we will explore the impact of bringing schedules from several models, and the viability of applying transfer-tuning across similar workload classes.
For future work, we will explore the impact of across-kernel interactions, the viability of applying transfer-tuning across similar kernel classes, and transfer-tuning across hardware devices.

%how transfer-tuning can be used to bootstrap auto-scheduling, as well as investigate its relevance to NAS.
% We leave exploration of these extensions and applications for future work.

% As future work we plan to
%*********************************************************************************************
\newpage
%%
%% The next two lines define the bibliography style to be used, and
%% the bibliography file.
\balance
\bibliographystyle{ACM-Reference-Format}
\bibliography{references}

% \onecolumn \begin{multicols}{2}

% \bibliographystyle{ACM-Reference-Format}
% \bibliography{references}

% \end{multicols}
\end{document}